%% file: main.tex
\documentclass{article}

\usepackage{microtype}
\usepackage{graphicx}
\usepackage{subcaption}
\usepackage{booktabs} 
\usepackage{pifont}
\usepackage{enumitem}
\usepackage{makecell}
\usepackage{multirow}
\usepackage{tabularx}
\usepackage{makecell}
\usepackage[table]{xcolor}
\usepackage{threeparttable}
\usepackage{adjustbox}
\usepackage{twemojis}
\usepackage{tcolorbox}
\usepackage{amssymb}
\tcbuselibrary{breakable}
\usepackage{fontawesome5}

\definecolor{moOchre}{HTML}{9C6F53} 
\definecolor{moGrey}{HTML}{50616D}  
\definecolor{moRed}{HTML}{9E5D66}

\definecolor{purple1}{RGB}{102, 0, 153} 
\definecolor{brown}{rgb}{0.6, 0.4, 0.2} 
\definecolor{bluegray}{HTML}{6B86A6}
\definecolor{sage}{HTML}{7E9B8E}
\definecolor{mauve}{HTML}{8D7A96}
\definecolor{warmtaupe}{HTML}{9A857A}
\definecolor{dustyrose}{HTML}{B58A8A}

\definecolor{ao}{rgb}{0.0,0.5,0.0}
\newcommand{\Yes}{{\color{ao}\ding{51}}}
\newcommand{\No}{{\color{red}\ding{55}}}

\usepackage{hyperref}
\usepackage{xcolor}  
\definecolor{iris}{rgb}{0.35, 0.31, 0.81}
\definecolor{amaranth}{rgb}{0.9, 0.17, 0.31}
\definecolor{ao}{rgb}{0.0, 0.5, 0.0}

\usepackage[preprint]{icml2026}

\usepackage{amsmath}
\usepackage{amssymb}
\usepackage{mathtools}
\usepackage{amsthm}

\usepackage[capitalize,noabbrev]{cleveref}

\theoremstyle{plain}

\theoremstyle{definition}

\theoremstyle{remark}

\usepackage[textsize=tiny]{todonotes}

\icmltitlerunning{AgentVista}

\begin{document}

\twocolumn[
  \icmltitle{\textsc{AgentVista}: Evaluating Multimodal Agents in Ultra-Challenging \\ Realistic Visual Scenarios}
  \icmlsetsymbol{equal}{*}

\begin{icmlauthorlist}
  \icmlauthor{Zhaochen Su}{hkust}
  \icmlauthor{Jincheng Gao}{hkust}
  \icmlauthor{Hangyu Guo}{hkust}
  \icmlauthor{Zhenhua Liu}{hkust}
  \icmlauthor{Lueyang Zhang}{hkust}
  \icmlauthor{Xinyu Geng}{hkust} \\
  \icmlauthor{Shijue Huang}{hkust}
  \icmlauthor{Peng Xia}{unc}
  \icmlauthor{Guanyu Jiang}{zju}
  \icmlauthor{Cheng Wang}{nus}
  \icmlauthor{Yue Zhang}{hkust}
  \icmlauthor{Yi R. (May) Fung}{hkust}
  \icmlauthor{Junxian He}{hkust}
  {\bf Website:} \href{https://agentvista-bench.github.io/}{\texttt{agentvista-bench.github.io}}
\quad \quad  \faGithub \ \  \href{https://github.com/hkust-nlp/AgentVista}{\texttt{github.com/hkust-nlp/AgentVista}}
\end{icmlauthorlist}

\icmlaffiliation{hkust}{Hong Kong University of Science and Technology}
\icmlaffiliation{zju}{Zhejiang University}
\icmlaffiliation{nus}{National University of Singapore}
\icmlaffiliation{unc}{University of North Carolina at Chapel Hill}
\icmlcorrespondingauthor{Zhaochen Su}{zsubf@connect.ust.hk}
\icmlcorrespondingauthor{Yi R. (May) Fung}{yrfung@ust.hk}
\icmlcorrespondingauthor{Junxian He}{junxianh@cse.ust.hk}
  \icmlkeywords{Machine Learning, ICML}

  \vskip 0.3in
]

\printAffiliationsAndNotice{}

\begin{abstract}
Real-world multimodal agents solve multi-step workflows grounded in visual evidence. For example, an agent can troubleshoot a device by linking a wiring photo to a schematic and validating the fix with online documentation, or plan a trip by interpreting a transit map and checking schedules under routing constraints. However, existing multimodal benchmarks mainly evaluate single-turn visual reasoning or specific tool skills, and they do not fully capture the realism, visual subtlety, and long-horizon tool use that practical agents require. We introduce \textsc{AgentVista}, a benchmark for generalist multimodal agents that spans 25 sub-domains across 7 categories, pairing realistic and detail-rich visual scenarios with natural hybrid tool use. Tasks require long-horizon tool interactions across modalities, including web search, image search, page navigation, and code-based operations for both image processing and general programming. Comprehensive evaluation of state-of-the-art models exposes significant gaps in their ability to carry out long-horizon multimodal tool use. Even the best model in our evaluation, \textsc{Gemini-3-Pro} with tools, achieves only 27.3\% overall accuracy, and hard instances can require more than 25 tool-calling turns. We expect \textsc{AgentVista} to accelerate the development of more capable and reliable multimodal agents for realistic and ultra-challenging problem solving.
\end{abstract}
\vspace{-15pt}
\section{Introduction}
Humans seamlessly integrate multi-sensory information to tackle complex real-world problems~\cite{stein2012new}. With the rapid evolution of AI agents~\cite{wang2024survey, comanici2025gemini,openai_o3_o4mini_system_card_2025,team2026kimi}, developing visual agentic intelligence becomes essential. For instance, an agent is expected to assist in shopping by scanning shelf products and retrieving nutritional information to satisfy user health constraints, or support troubleshooting by linking malfunction photos with schematic diagrams to diagnose specific faults. However, a major challenge in developing such multimodal agents is the absence of a benchmark based on realistic scenarios that covers the diversity and complexity of long-horizon tool interactions across different modalities, which limits reliable evaluation of agent capabilities in open domains~\cite{xie2024osworld,li2025tool}.
\begin{figure}[t]
    \centering
    \includegraphics[width=0.95\linewidth]{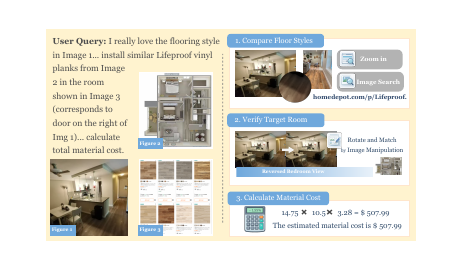}
    \caption{A representative \textsc{AgentVista} task grounded in a real home-renovation scenario. The agent needs to match flooring styles across images, verify the target room, retrieve product specifications, and compute final cost via interleaved tool use.}
    \label{fig:example}
\vspace{-15pt}
\end{figure}

\begin{figure*}[t]
    \centering
    \includegraphics[width=0.95\textwidth]{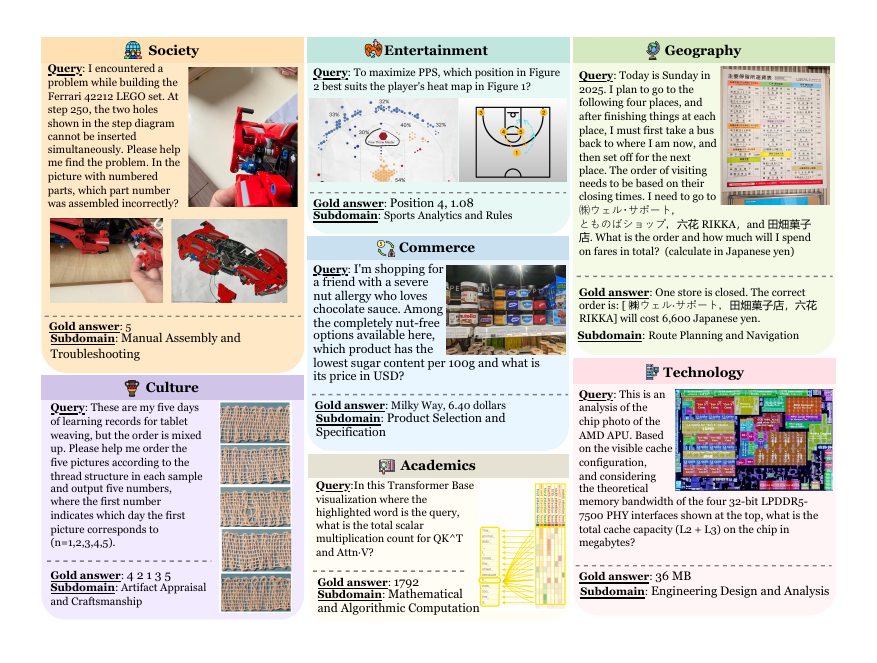}
     \caption{Sampled \textsc{AgentVista} examples from each domain. Each query is grounded in complex, real-world visual scenes and is designed to elicit agentic tool use with multi-step reasoning toward a unique, verifiable answer.}
    \vspace{-5pt}
    \label{fig:data}

\end{figure*}

Traditional multimodal benchmarks~\cite{Antol_2015_ICCV, Hudson_2019_CVPR, Yue_2024_CVPR, wang2024charxiv, scaleai_vista_2025} focus on assessing visual perception and complex reasoning capabilities. Recently, a growing number of benchmarks have emerged to evaluate multimodal agentic behaviors~\citep{ma2024m,li2025tir,ashraf2025agent,guo2025beyond,tao2025mmsearch,geng2025webwatcher}. However, these evaluations typically present two main gaps: 
\ding{182} \textbf{Capability-Specific Evaluation}: They typically emphasize particular capabilities, focusing on skills such as visual manipulation~\cite{hrbench,lai2025mini}, web browsing~\cite{li2025mmbrowsecomp,tao2025mmsearch}, or code generation~\cite{yang2024swebenchmm}. This narrow focus makes it difficult to evaluate generalist agents that must combine multiple skills and remain reliable in long-horizon workflows. 
\ding{183} \textbf{Trade-off between Realism and Difficulty}: Practical agent tasks are difficult because they combine cluttered visual evidence with long-horizon tool use under constraints. 
Yet many benchmarks increase difficulty by simplifying the visual state or by relying on tool patterns that deviate from everyday workflows, which can shift the bottleneck away from realistic grounding and interaction. For example, VisualToolBench pre-processes the input images to facilitate specific visual operations~\cite{guo2025beyond}. While this design is effective for evaluating visual manipulation, it also shifts the problem from reasoning over natural visual states to operating on curated inputs.

\input{table/comparison_datasets}

To address these gaps, we introduce \textsc{AgentVista}, a benchmark designed to evaluate generalist multimodal agents on diverse, realistic, and challenging tasks. \textsc{AgentVista} contains 209 tasks spanning 25 sub-domains across 7 categories, including commerce, geography, society, technology, entertainment, culture, and academics, and grounds each query in detail-rich visual states such as daily photos, screenshots, and technical diagrams, with both single-image and multi-image inputs. Each query is manually authored to reflect authentic user intent and is subjected to strict quality control, where every instance is carefully reviewed to ensure mandatory visual dependence and a unique, verifiable answer. Every task requires long-horizon interaction with interleaved tools, where the agent repeatedly grounds visual cues, retrieves external information, and verifies intermediate decisions. Table~\ref{tab:agent_benchmark_comparison} summarizes the key differences between \textsc{AgentVista} and representative agentic multimodal benchmarks. Figure~\ref{fig:example} shows a representative example from \textsc{AgentVista} motivated by a real home renovation need: the agent need to match flooring styles across scenes, verify the target room with image-based checks, retrieve product specifications online, and compute a deterministic final cost from the room size and packaging information.

\textsc{AgentVista} is evaluated in a controlled yet practical setting, we adopt four widely used tools that cover the core interaction patterns of real-world multimodal agents, including web search, image search, page navigation, and code-based operations for both image processing and general programming. Our experiments on representative open-source and commercial MLLMs show that \textsc{AgentVista} remains far from being solved, leaving substantial room for improvement. Even the best performance in our evaluation, \textsc{Gemini-3-Pro}, achieves only 27.3\% overall accuracy. Further error analysis shows that many failures start with visual misidentification and then lead to wrong retrieval and unreliable tool use over many steps. To facilitate future research, we will release both the \textsc{AgentVista} benchmark and a lightweight yet general agent framework to facilitate reproducible evaluation and accelerate progress on long-horizon multimodal tool use.

\section{The \textsc{AgentVista}}

\begin{figure*}[t]
    \centering
    \includegraphics[width=1\textwidth]{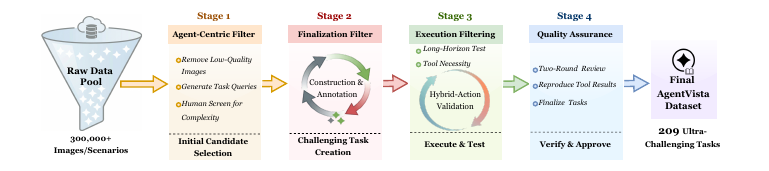}
    \vspace{-5pt}
    \caption{Overview of the \textsc{AgentVista} dataset construction pipeline, consisting of agent-centric filtering, expert finalization, execution filtering, and two-round verification to produce realistic and ultra-challenging multimodal agent tasks.}
    \vspace{-10pt}
    \label{fig:compare_paradigms}

\end{figure*}

\begin{figure}[t]
    \centering
    \includegraphics[width=0.8\linewidth]{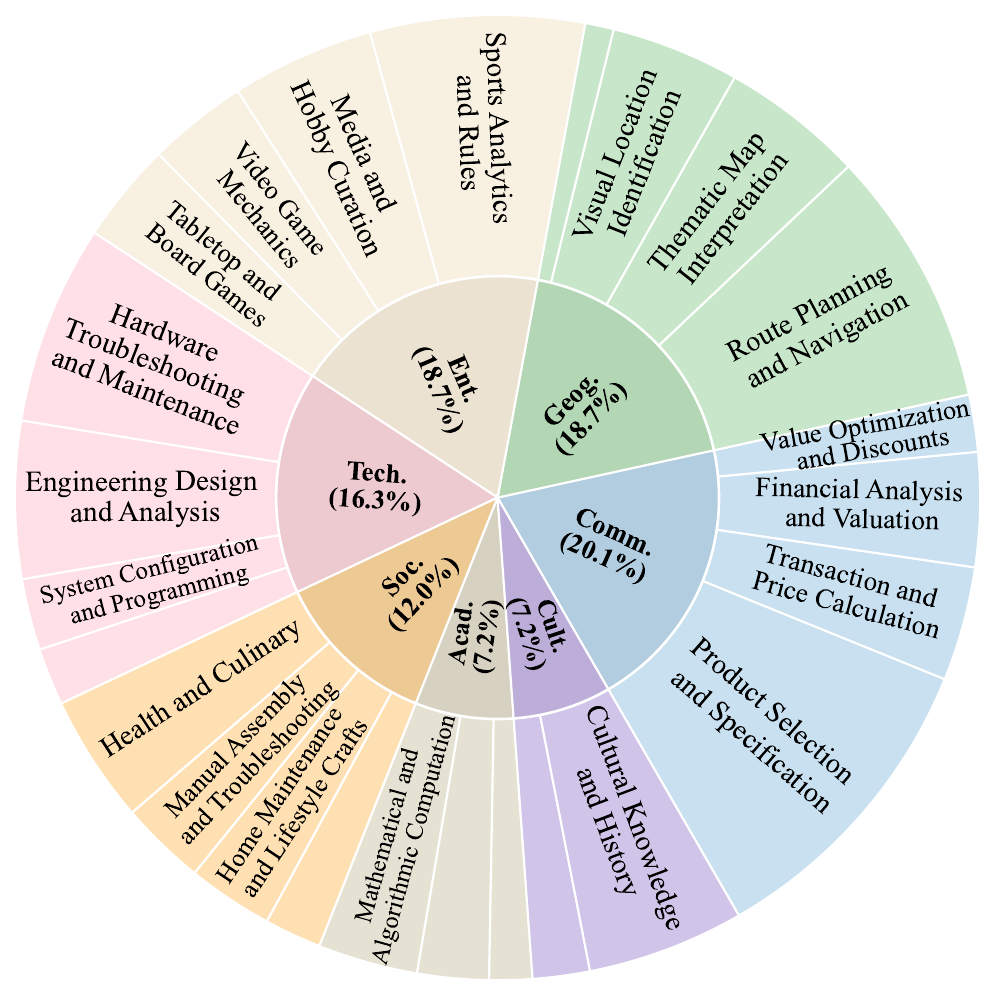}
    \vspace{-5pt}
    \caption{The categorization of \textsc{AgentVista}. The benchmark spans 7 major categories and 25 sub-domains, covering a broad range of realistic and challenging multimodal agent scenarios.
    Category abbreviations:
    \textbf{\textsc{Comm.}} (Commerce), \textbf{\textsc{Geog.}} (Geography), \textbf{\textsc{Ent.}} (Entertainment),
    \textbf{\textsc{Tech.}} (Technology), \textbf{\textsc{Soc.}} (Society), \textbf{\textsc{Acad.}} (Academics),
    and \textbf{\textsc{Cult.}} (Culture).}
    \vspace{-14pt}
    \label{fig:data_stastics}
\end{figure}

\subsection{Overview of \textsc{AgentVista}}
We introduce \textsc{AgentVista}, a benchmark for evaluating generalist multimodal agents on realistic and ultra-challenging tasks. \textsc{AgentVista} focuses on realistic user requests that are still hard in practice and require long-horizon tool use grounded in visual evidence. \textsc{AgentVista} contains 209 tasks spanning 25 sub-domains across 7 categories: Technology, Commerce, Geography, Entertainment, Society, Academics, and Culture. The domain distribution and dataset composition are summarized in Table~\ref{tab:statistics} and Figure~\ref{fig:data_stastics}. As shown in Figure~\ref{fig:data}, tasks are built from authentic user needs and require multi-step reasoning with tool use. For example, an agent may need to read key constraints from a photo or screenshot, retrieve missing details from external resources, and then combine multiple pieces of evidence to produce the final answer. This includes diagnosing a hardware issue by matching visible components to technical documentation, selecting a product that satisfies allergy and nutrition constraints by comparing labels with online specifications, and planning a route under time and transit limits by reading schedules from images and verifying them with web search. To enable robust and scalable evaluation, each instance is paired with a clear and deterministic ground truth answer, typically a short phrase or a numeric value.

\input{table/statistics}
\subsection{Data Construction}

\subsubsection{Core Design Principles}
\vspace{-5pt}
\label{sec:design_principles}
We design \textsc{AgentVista} based on three principles:
\vspace{-10pt}
\begin{itemize}[leftmargin=*]
\setlength{\itemsep}{0pt}
\item \textbf{Vision-centric tasks with realistic images.} Each task requires obtaining the key evidence from the visual input. The images are real and contain visual details to support visual understanding, such as small but important cues, multiple related objects, or subtle differences across views. The query avoids stating the key information in text and avoids questions that can be answered by a keyword search. These constraints ensure that solving the task relies on understanding and comparison of visual details, rather than on textual shortcuts.

\item \textbf{Natural interleaved hybrid tool use.} Each task requires using different tool types together, and the interaction must include interleaved tool calls across at least two tool categories. The intended solution should mix visual tools and text-based tools, such as using image search or image processing to gather visual evidence, then using web search or page navigation to retrieve needed facts, and finally combining the evidence to reach the answer. Tool use must follow natural and real-world workflows. Each tool call should be necessary for solving the task, rather than added only to make the interaction longer. To keep tasks realistic and challenging, we favor instances that require grounding tool outputs in the visual input under explicit constraints.

\item \textbf{Easy to verify and stable over time.} Following recent evaluation protocols~\cite{li2025mmbrowsecomp, wei2025browsecomp}, each task has a concise target answer in a fixed format, such as a number, an entity name, or a short description. This design makes the evaluation process simple and accurate, similar to math tasks. Additionally, we address the issue of information changing over time. Annotators verify facts against reliable sources. When necessary, we include specific time constraints in the question to ensure the ground truth remains valid.

\end{itemize}

\subsubsection{Dataset Creation Pipeline}
We build \textsc{AgentVista} from 300k+ real images and real user needs collected from public model arenas, annotator-captured daily scenarios, and private community forums, with details in Appendix~\ref{sec:data_source}. The dataset construction pipeline is shown in Figure~\ref{fig:compare_paradigms}.

\vspace{-10pt}
\paragraph{Stage 1: Agent-centric filtering.}
We start with model-assisted mining and filtering to identify candidate initial states that reflect realistic daily workflows. We first use \textsc{Claude-Opus-4} to filter the raw image pool by removing cases with limited visual information or weak agentic potential, such as pure OCR screenshots, single-object landmark photos, and images that can be solved without meaningful visual reasoning. We provide \textsc{Claude-Opus-4} with our tool schema and ask it to propose an initial task query that is compatible with the available tools and has a verifiable answer format. The proposed query serves as a candidate starting point for downstream curation. We then apply human screening to retain only images with sufficiently rich visual evidence and queries that support a natural task formulation with hybrid tool use. To avoid simple cases, we prioritize candidates with non-trivial constraints and keep only those that naturally require multi-step reasoning rather than a single direct lookup.

\vspace{-5pt}
\paragraph{Stage 2: Expert finalization.}
We recruit and train expert annotators on the project scope, taxonomy, and quality requirements, and ask them to finalize each candidate produced in Stage~1. Starting from the image and the initial query, annotators rewrite the query into a realistic user request while keeping it self-contained and vision-centric. Realism is enforced by preserving the original visual state and intent, and by expressing constraints in the way users typically specify them, such as time, budget, compatibility, and safety requirements. To make tasks ultra-challenging in a natural way, annotators select cases where the answer depends on fine-grained visual cues and cannot be obtained by a single direct lookup. They ensure that solving the task requires combining visual evidence with information gathered from tools, and that the process includes necessary interleaving across tool types. Annotators then produce a deterministic target answer and record the key evidence and tool steps used to obtain it, which enables later checking.

\vspace{-5pt}
\paragraph{Stage 3: Execution filtering.}
We validate each instance by executing the candidate task in our tool environment and checking that the annotated answer is supported by reproducible tool outputs. During this process, we run \textsc{Gemini-3-Flash} in the same tool environment to screen for tool-use diversity, and we retain only tasks that require interleaved calls across at least two categories. Furthermore, we run \textsc{Gemini-2.5-Pro} with tool access disabled and remove samples that can be solved from the prompt alone.

\vspace{-5pt}
\paragraph{Stage 4: Two-round verification.}
Finally, we conduct two rounds of verification. The first round removes instances with insufficient visual evidence, weak visual dependency, or questionable answer validity. In the second round, a separate group re-checks each instance by following the evidence and tool steps recorded by annotators, and confirms that the final answer is supported by the visual cues and the tool outputs. Instances with unclear evidence, unstable answers, or unrealistic workflows are removed. The remaining instances form the final \textsc{AgentVista} benchmark.

\input{table/main_table}
\vspace{-5pt}
\paragraph{Filtering statistics.}
We begin with 300k+ candidate images. Stage~1 uses model-assisted filtering and human screening to select 568 potential initial states, 0.19\% of the raw pool. Stage~2 expert finalization yields 315 tasks after rewriting the initial queries into realistic user requests and adding deterministic target answers. Stage~3 execution filtering retains 241 tasks by validating reproducible tool outputs, enforcing interleaved calls across at least two tool categories, and removing tasks solvable when tool access is disabled. Stage~4 two-round verification selects the final 209 tasks by re-checking visual evidence, recorded tool steps, and answer validity. On average, constructing a single instance takes about 4 hours, and expert annotators take about 30 minutes to solve an instance.

\vspace{-5pt}

\subsubsection{Tool environment}
\textsc{AgentVista} supports a compact set of tools that cover common multimodal agent workflows. Models can call \texttt{web\_search} to retrieve web pages, \texttt{visit} to open and navigate a page, and \texttt{image\_search} to locate images when a query requires external visual references. We also provide \texttt{code\_interpreter}, which supports both programming and image processing. It enables arithmetic and parsing, structured extraction, and operations such as cropping, resizing, measuring, and comparing visual regions when needed. All tools are exposed with detailed descriptions and structured inputs and outputs, so the model can decide when to call a tool and how to use the returned results. Detailed tool definitions are provided in Appendix~\ref{sec:tool_definition_preface}.

\vspace{-5pt}

\section{Experiments}
\subsection{Experimental Setup}
\paragraph{Models.}
We evaluate a broad set of frontier multimodal models that are commonly used as generalist agents. Specifically, we test \textsc{GPT-4.1}~\citep{openai2025gpt41}, \textsc{o3}, \textsc{o4-mini}~\citep{oai2025o3o4mini}, \textsc{GPT-5}~\citep{openai2024gpt5}, \textsc{GPT-5.1}~\citep{openai2025gpt51}, \textsc{GPT-5.2}~\citep{openai2025gpt52}, \textsc{Gemini-3-Flash}~\citep{deepmind2025gemini3flash}, \textsc{Gemini-3-Pro}~\citep{deepmind2025gemini3pro}, \textsc{Grok-4}~\citep{xai2025grok4}, \textsc{Claude-Sonnet-4}~\citep{anthropic2025claudesonnet4}, \textsc{Claude-Opus-4.1}~\citep{anthropic2025claudeopus41}, \textsc{Claude-Sonnet-4.5}~\citep{claude45sonnet}, and \textsc{Qwen3-VL-235B-A22B}~\citep{bai2025qwen3vl}.

\vspace{-10pt}

\paragraph{Evaluation Setup.} 
For all experiments, we use a temperature of 0.6 and cap the tool interaction budget at 30 turns for every model. Since \textsc{AgentVista} provides concise target answers in deterministic formats, evaluation reduces to verifying the final answer. We use \textsc{GPT-4.1} as a fixed judge model to assess whether a model's final response matches the annotated ground truth under the required format. We report accuracy as the evaluation metric.


\subsection{Main Results}
\vspace{-5pt}
We report the overall performance in Table~\ref{main-table}. We make the below  three observations.
\vspace{-10pt}
\paragraph{\textsc{AgentVista} is ultra-challenging.}
The results show that \textsc{AgentVista} remains difficult for current multimodal agents. Even the best-performing model, \textsc{Gemini-3-Pro}, achieves 27.27\% overall accuracy, indicating substantial headroom. Performance is also low for a large portion of models: 4 out of 14 models score below 15\% overall accuracy. These results suggest that agents still have significant room for improvement in complex long-horizon settings that require multi-step tool use grounded in real visual evidence. The average number of turns further reflects this difficulty. For example, \textsc{GPT-5.2} uses 13.85 turns on average, and 5 out of 14 models exceed 10 turns on average, indicating that many tasks require extended multi-step interactions rather than a short tool sequence. We also observe a sizable gap between the open-source model \textsc{Qwen3-VL-235B} and the closed-source models, suggesting substantial room for open-source multimodal agents. We report additional open-source baselines in Appendix~\ref{apx:open_source}. We further analyze common failure patterns in Section~\ref{sec:error_analysis}.

\vspace{-15pt}
\paragraph{Domain strengths differ across model families.}
Performance varies noticeably across categories, revealing complementary strengths among model series. The GPT-5 family shows strong coverage on practical categories, with \textsc{GPT-5.2} performing best on \textsc{Technology} and tying for the best score on \textsc{Entertainment}, while \textsc{GPT-5} and \textsc{GPT-5.1} lead \textsc{Commerce}. The Gemini series is strongest overall: \textsc{Gemini-3-Pro} achieves the highest overall accuracy, leads \textsc{Geography}, and performs competitively on \textsc{Society} and \textsc{Culture}. Claude models are comparatively stronger on categories that emphasize careful reading and constraint following, with their best results appearing in \textsc{Technology} and \textsc{Geography}. Overall, these results suggest that current agents do not yet provide uniform competence across domains, and improving broad, consistent performance across realistic long-horizon tasks remains an open challenge.

\vspace{-8pt}
\paragraph{Multi-image inputs are not uniformly harder than single image inputs.}
For nearly all evaluated models, accuracy with multi-image inputs is higher than with single-image inputs. The gain is especially large for \textsc{Gemini-3-Pro}, which improves from 23.68\% under single-image input to 36.84\% under multi-image input. This pattern matches how our multi-image instances are constructed. Additional views often provide complementary evidence, reduce ambiguity, and reveal details that are missing in a single shot, which can make grounding and downstream retrieval more reliable. While multi-image inputs still require cross-image alignment, the results suggest that the main bottleneck remains long-horizon tool use and constraint tracking, rather than the presence of multiple images itself.
\vspace{-6pt}

\section{Further Analysis}
\vspace{-5pt}
\begin{figure}[t]
    \centering
    \includegraphics[width=\linewidth]{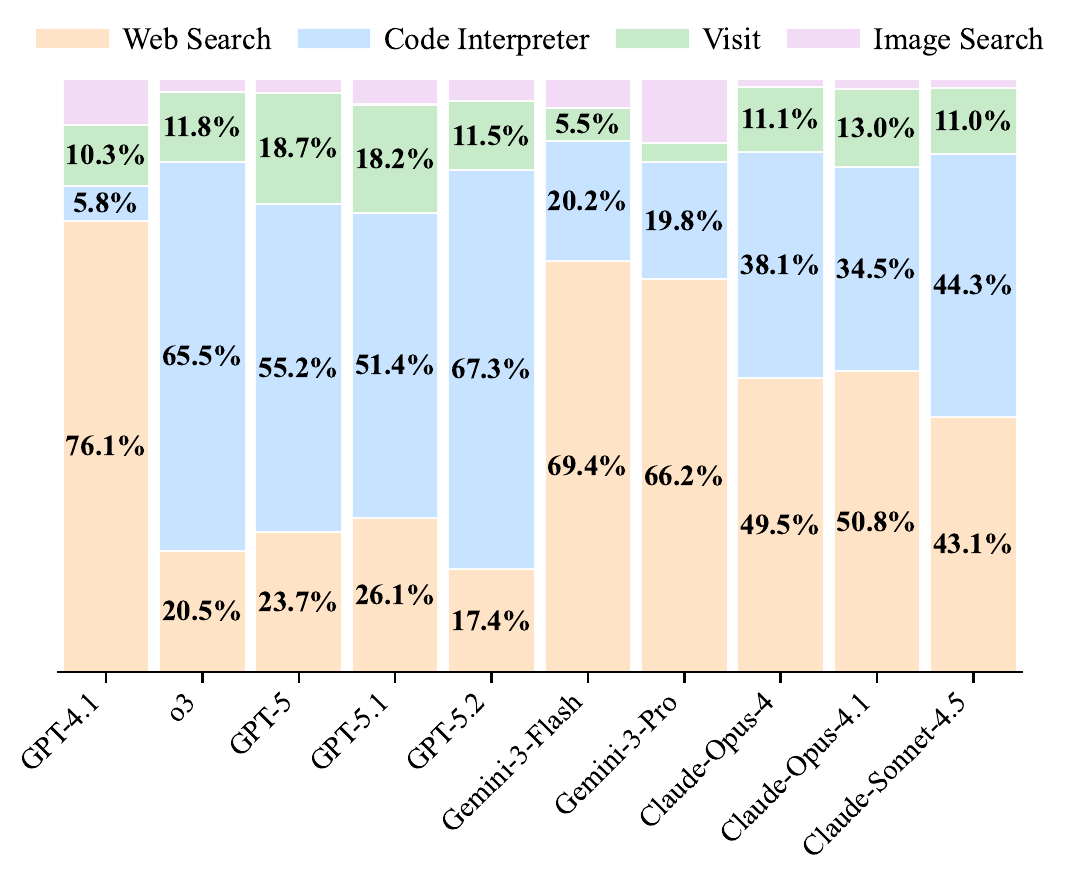}
    \vspace{-20pt}
    \caption{Tool-use distribution across models. GPT models rely more on the code interpreter, while Gemini and Claude models use web search most frequently.}
    \label{fig:tool_distribution}
    \vspace{-15pt}

\end{figure}

\subsection{Tool Distribution Analysis}
\begin{figure*}[t]
    \centering
    \includegraphics[width=\textwidth]{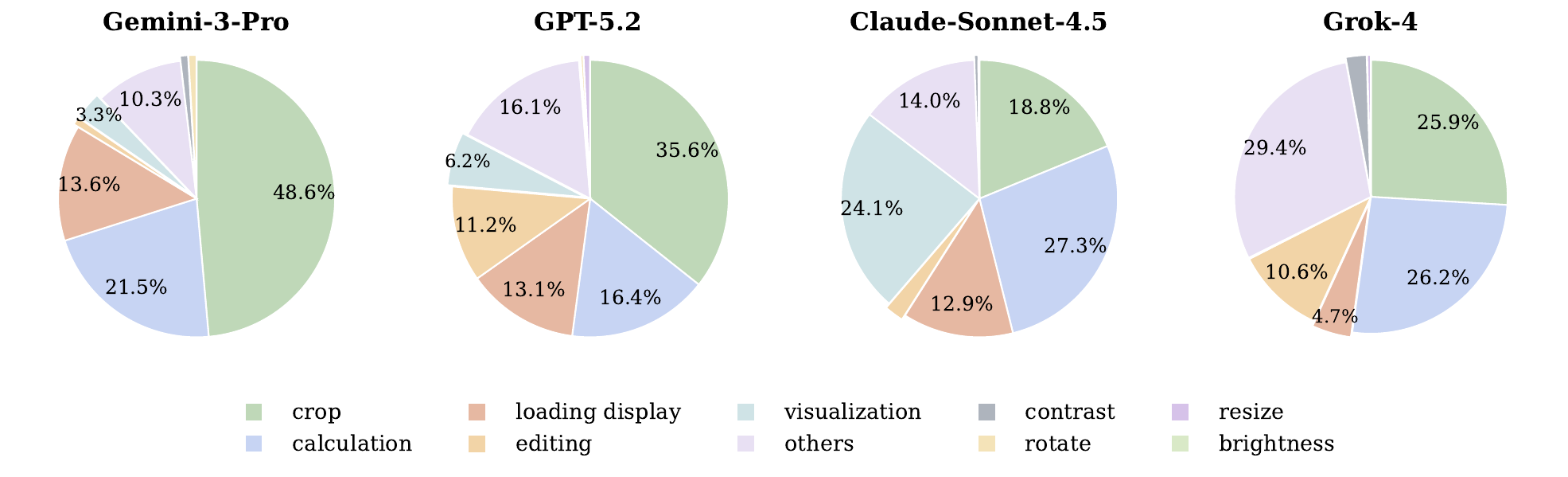}
    \caption{Image manipulation operation distribution of code interpreter calls across four multimodal models. Tool usages are automatically categorized into image-editing and analysis-related types. Across models, \textit{crop} is the most frequent operation, suggesting that many interactions rely on localized visual grounding before further reasoning.}
    \label{fig:codeintepreter_type}
\end{figure*}

\vspace{-5pt}
In this section, we analyze the distribution of tool calls across models. As shown in Figure~\ref{fig:tool_distribution}, the \textsc{GPT-5} series relies most heavily on the code interpreter. We further break down code interpreter calls by operation type in Figure~\ref{fig:codeintepreter_type}. The results suggest that these models more often perform image-centric operations during problem solving, such as zooming in, cropping, resizing, measuring regions, and carrying out structured extraction or calculations.  Across the inspected models, \textit{crop} is the most frequent operation, indicating that many trajectories depend on localized visual grounding before proceeding to retrieval or computation. Second, the \textsc{Gemini} and \textsc{Claude} series call web search most often, indicating a stronger preference for retrieval-driven workflows. Across all models, image search is used less frequently than the other tools. In the next tool ablation study, we quantify how each tool contributes to performance and how accuracy changes when a tool is removed.

\vspace{-5pt}
\subsection{Tool Ablation Study}
In this section, we ablate tool access to quantify how each tool modality contributes to performance.
\vspace{-15pt}
\paragraph{Experimental setup.}
We evaluate three settings with prompts lightly adapted to reflect the available capabilities, while keeping the evaluation protocol and inference hyperparameters fixed. \ding{182} \textbf{Vision only}: the agent has access only to a visual manipulation environment, enabling image processing operations for inspection and transformation, but no external retrieval. \ding{183} \textbf{Search only}: the agent can retrieve external evidence through both image-based and text-based search, and can read retrieved webpages, but cannot perform tool-based visual manipulation or programmatic verification. \ding{184} \textbf{No tool}: the agent relies purely on direct generation without any tool assistance.

\begin{figure}[t]
    \centering
    \includegraphics[width=\linewidth]{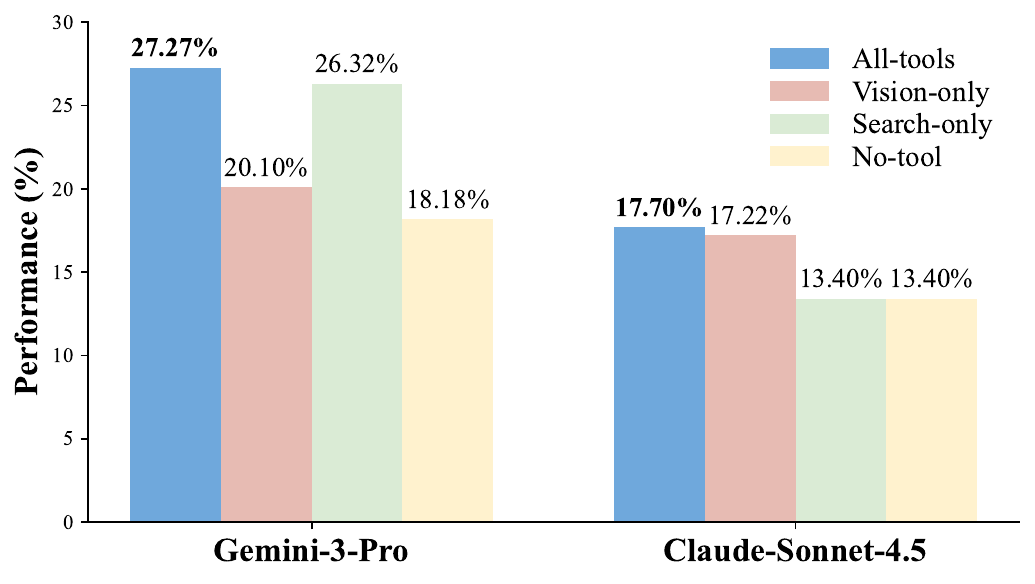}
    \vspace{-5pt}
    \caption{Tool ablation on \textsc{Gemini-3-Pro} and \textsc{Claude-Sonnet-4.5}. Both models perform best with the full tool suite, highlighting the importance of combining visual manipulation and retrieval.}
    \label{fig:ablation}
    \vspace{-10pt}
\end{figure}

\vspace{-10pt}
\paragraph{Key findings.}
Figure~\ref{fig:ablation} shows that using the full tool suite yields the best performance for both models, confirming that \textsc{AgentVista} rewards hybrid workflows that combine visual manipulation and retrieval. For \textsc{Gemini-3-Pro}, the full tool setting reaches 27.27\% accuracy, higher than the vision-only setting at 20.10\% and the no-tool setting at 18.18\%. For \textsc{Claude-Sonnet-4.5}, the full tool setting achieves 17.70\%, slightly above the vision-only setting at 17.22\%, while the search-only and no-tool settings both drop to 13.40\%. We also find that the role of retrieval differs across models. For \textsc{Gemini-3-Pro}, the search-only setting reaches 26.32\%, close to the full tool setting. This suggests that its strong visual perception enables it to extract reliable cues from images and benefit primarily from retrieval and page navigation, while visual manipulation mainly supports inspection and verification. In contrast, \textsc{Claude-Sonnet-4.5} relies more on visual manipulation than retrieval, since the vision-only setting remains close to the full tool setting, whereas the search-only setting degrades substantially.

\vspace{-10pt}

\begin{figure*}[t]
    \centering
    \includegraphics[width=\textwidth]{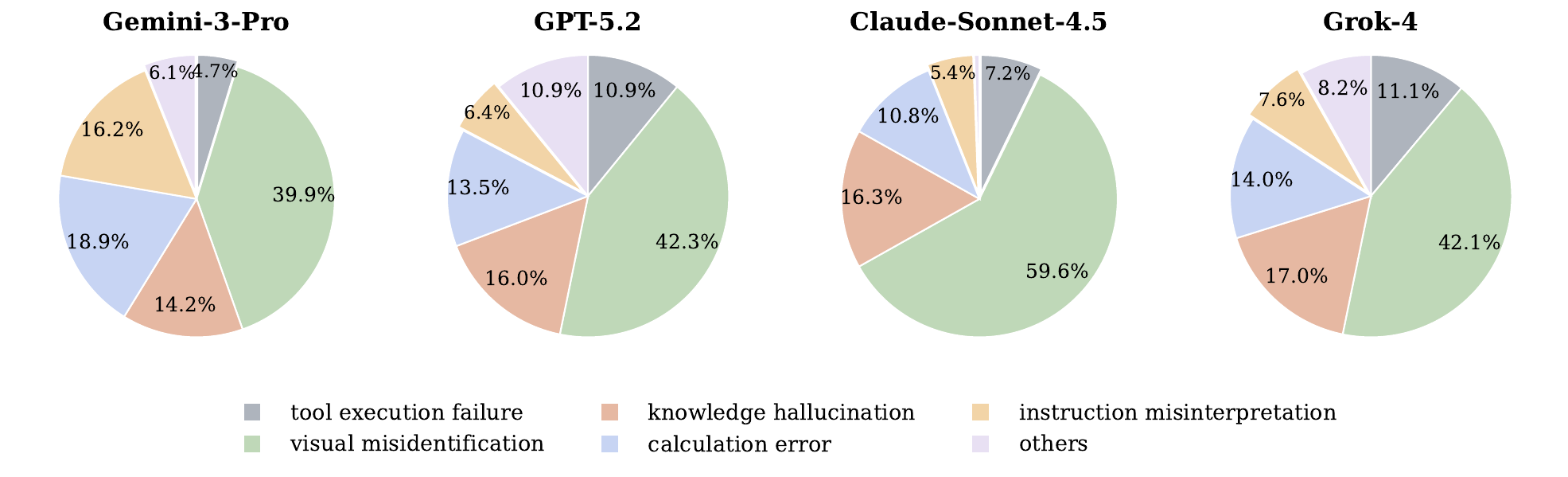}
    \caption{Error category distribution on \textsc{AgentVista} across four multimodal models. Error types are automatically labeled by \textsc{Gemini-3-Flash} based on model trajectories. Across all models, \textit{visual misidentification} is the dominant failure mode, indicating that many errors originate from incorrect grounding on fine-grained visual evidence.}
    \label{fig:error_category}
    \vspace{-10pt}
\end{figure*}

\subsection{Error Analysis}
\vspace{-5pt}
\label{sec:error_analysis}
To understand the main bottlenecks on \textsc{AgentVista}, we analyze failures from four representative models. For each incorrect case, we assign an error label, including tool execution failure, visual misidentification, knowledge hallucination, calculation error, instruction misinterpretation, and others. The labels are generated by \textsc{Gemini-3-Flash} based on the model trajectories, and the distributions are shown in Figure~\ref{fig:error_category}. Detailed definitions for each error type are provided in Appendix~\ref{apx:error}. Figure~\ref{fig:error_category} shows a clear trend that visual misidentification is the main failure mode across all models. This aligns with the design of \textsc{AgentVista}, where tasks are grounded in realistic and cluttered visual states and often depend on small but critical details. From bad cases, we find that frontier agents can often zoom in to the relevant region, but they still fail when the image is blurry or the key cue is visually subtle. Knowledge hallucination is the second most common error type, which also matches our benchmark design. Many tasks require applying diverse world knowledge to long-horizon tool interactions, and current models still struggle to resolve long-tail facts reliably even with web search. We include representative good and bad cases with detailed explanations in Appendix~\ref{apx:case_study}. Overall, these results suggest that \textsc{AgentVista} can expose practical weaknesses in both fine-grained visual understanding and knowledge-grounded reasoning under realistic tool use.

\subsection{Test Time Scaling}
\begin{table}[t]
\small
\centering
\caption{Test-time scaling results under different sampling budgets $K$ on \textsc{Gemini-3-Flash}. We report Random1@$K$ as a lower bound, Best-of-$K$ (BoN@$K$) selected by a reward model, and Pass@$K$ as an upper bound. All values are accuracies in \%.}
\label{tab:tts_scaling}
\resizebox{1.0\linewidth}{!}{%
\begin{tabular}{lccccc}
\toprule
\textsc{\textbf{Setting}} & \textbf{\textsc{$K{=}1$}} & \textbf{\textsc{$K{=}2$}} & \textbf{\textsc{$K{=}4$}} & \textbf{\textsc{$K{=}8$}} & \textbf{\textsc{$K{=}16$}} \\
\midrule
\textbf{\textsc{Random1@$K$}} & 21.05 & 19.11 & 18.23 & 17.09 & 18.05 \\
\textbf{\textsc{BoN@$K$}}         & 21.05 & 24.88 & 26.32 & 28.23 & 30.62 \\
\textbf{\textsc{Pass@$K$}}        & 21.05 & 26.07 & 34.22 & 42.59 & 51.67 \\
\bottomrule
\end{tabular}}
\vspace{-10pt}
\end{table}

\label{sec:tts}
To study whether additional sampling at test time can improve performance on \textsc{AgentVista}, we evaluate test-time scaling on \textsc{Gemini-3-Flash}. We generate $K$ independent solutions per instance and use \textsc{Gemini-3-Flash} as the reward model to select a final answer when selection is required. We follow the same evaluation protocol as in prior experiments. Table~\ref{tab:tts_scaling} reports three settings: \textsc{Random1@$K$}, which randomly selects one of the $K$ samples as a lower bound, Best-of-$K$ (\textsc{BoN@$K$}), which selects the highest-scoring sample under the reward model, and \textsc{Pass@$K$}, which measures whether at least one of the $K$ samples is correct as an upper bound.
\vspace{-5pt}
\paragraph{Key findings.}
Table~\ref{tab:tts_scaling} shows that test-time scaling consistently improves performance. Under \textsc{BoN}, accuracy increases from 21.05\% at $K{=}1$ to 30.62\% at $K{=}16$. The upper bound rises even more, with \textsc{Pass@$K$} increasing from 21.05\% at $K{=}1$ to 51.67\% at $K{=}16$. In contrast, \textsc{Random1@$K$} remains low and does not improve with larger $K$, indicating that gains mainly come from better selection rather than sampling alone. Despite these improvements, scaling alone is not sufficient to solve \textsc{AgentVista}. Even at $K{=}16$, \textsc{BoN} reaches only 30.62\%, while {Pass@$16$} is 51.67\%. This gap indicates substantial room for reinforcement learning or other optimization methods that can better close the gap between selection and the achievable upper bound, and more broadly highlights the need for stronger long-horizon tool use and more reliable visual grounding.

\section{Related Work}
\subsection{Multimodal Agents and Tool Use}
Recent years have witnessed rapid progress in large multimodal models that combine visual perception with language-based reasoning~\cite{peng2023kosmos, liu2023visual, zhu2023minigpt, li2023blip}. A key step toward practical multimodal agents is to couple these models with tools so they can inspect visual evidence, verify intermediate hypotheses, and refine solutions over multiple steps. OpenAI o3 and o4-mini follow this direction by manipulating user-provided images during reasoning through operations such as cropping, zooming, and rotation, and coordinating these visual operations with other tools when needed~\cite{openai_o3_o4mini_system_card_2025}. This paradigm has inspired open systems that study tool-driven multimodal reasoning and long-horizon interaction~\cite{su2025openthinkimg, su2025thinking}. Recent work also explores stronger training signals for repeated grounding, such as reinforcement learning for interleaved perception and reasoning~\cite{zheng2025deepeyes}, and extends multimodal agents with web and code tools for mixed tool use in realistic settings~\cite{hong2025deepeyesv2, geng2025webwatcher}. Despite this progress, there is still no benchmark that evaluates generalist multimodal agents on realistic, ultra-challenging tasks. \textsc{AgentVista} fills this gap by focusing on long-horizon, interleaved tool use grounded in real visual inputs.

\vspace{-5pt}

\subsection{Multimodal Agent Benchmarks}
Early multimodal benchmarks mainly evaluate perception and visual reasoning in static question answering, where models respond from a fixed image and text context without interaction~\cite{Antol_2015_ICCV,Hudson_2019_CVPR,lu2023mathvista,yue2024mmmu,wang2024charxiv}. While useful, they do not test whether an agent can choose actions, call tools, and verify intermediate results. Recent agent benchmarks add tool use, including multi-step planning~\cite{ma2024m}, web browsing and search~\cite{li2025mmbrowsecomp,tao2025mmsearch}, and tool-assisted visual reasoning and active perception~\cite{wu2024v,lai2025mini,li2025tir,ashraf2025agent}. More recent works further move toward interleaved tool settings, but the visual evidence is often relatively clean or lightweight, which makes perception less demanding, and the resulting tool trajectories tend to be shorter and less diverse~\cite{guo2025beyond, hong2025deepeyesv2, chen2026mindwatchersmartermultimodaltoolintegrated}. \textsc{AgentVista} addresses this gap by emphasizing realistic visual inputs and long-horizon workflows that require repeated visual checking and interleaved use of multiple tool types.

\vspace{-5pt}

\section{Conclusion}
We introduce \textsc{AgentVista}, a benchmark for evaluating generalist multimodal agents on realistic, ultra-challenging tasks that require long-horizon, interleaved tool use grounded in visual evidence. \textsc{AgentVista} contains 209 tasks spanning 25 sub-domains across 7 categories, with strict quality control to ensure vision-centric queries and unique, verifiable answers. Experiments across frontier models show that \textsc{AgentVista} is far from solved: even the best-performing model, \textsc{Gemini-3-Pro}, reaches only 27.3\% overall accuracy. The benchmark also elicits long interaction trajectories, with models such as \textsc{GPT-5.2} averaging 13.85 tool turns per task, indicating substantial complexity beyond short tool chains. Further analysis highlights visual grounding and long-horizon tool use as key bottlenecks for current multimodal agents. We hope \textsc{AgentVista} provides a practical benchmark for tracking progress and motivates the development of multimodal agents that can solve complex, multi-step real-world tasks more reliably.

\section*{Impact Statement}
This work introduces \textsc{AgentVista}, a benchmark for evaluating generalist multimodal agents on realistic, ultra-challenging tasks that require long-horizon tool use grounded in real visual inputs. By using concise, verifiable answers and a controlled tool environment, \textsc{AgentVista} enables reproducible comparisons and helps identify key bottlenecks in visual grounding, constraint tracking, and tool reliability. Improved multimodal agents could benefit practical applications such as shopping assistance, travel planning, and troubleshooting from user photos, where agents must combine visual evidence with online information and computation. At the same time, stronger agents may increase risks of privacy leakage from user-provided images and overconfident but incorrect outputs in real deployments. We mitigate these concerns by filtering and rewriting tasks to avoid personal identifiers when applicable, and by emphasizing short answers that encourage checkable evaluation rather than persuasive free-form text.

Benchmark construction can also reflect biases from source data and annotator decisions, which may affect coverage across domains and scenarios. We hope \textsc{AgentVista} supports future work on more robust and responsible multimodal agents by providing a shared evaluation target for realistic, long-horizon tool use.

\nocite{langley00}

\bibliography{example_paper}
\bibliographystyle{icml2026}

\newpage
\appendix
\onecolumn

\section{\textsc{AgentVista} Details}
\subsection{Dataset Taxonomy of \textsc{AgentVista}}
\label{sec:taxonomy}
\textsc{AgentVista} covers seven major categories: (1) \textbf{Technology}, which includes hardware troubleshooting, engineering analysis, and system configuration grounded in real photos, screenshots, and diagrams; (2) \textbf{Commerce}, which includes product selection, pricing and budget calculation, and finance-related reasoning under practical constraints; (3) \textbf{Geography}, which includes route planning, map interpretation, location identification, and spatial calculations; (4) \textbf{Entertainment}, which includes sports analytics, media and hobby curation, and game-related reasoning; (5) \textbf{Society}, which includes everyday tasks such as health and culinary decisions, home maintenance, manual assembly troubleshooting, and plant care; (6) \textbf{Academics}, which includes mathematical computation, scientific identification, and data analysis; and (7) \textbf{Culture}, which includes cultural knowledge, history-related understanding, and artifact appraisal grounded in visual evidence.

\subsection{Data Sources}
\label{sec:data_source}
All \textsc{AgentVista} instances are grounded in real images and real user needs. Across all sources, we apply a unified set of criteria. We retain only images with sufficient visual detail to support non-trivial reasoning, and we exclude cases where the solution can be obtained by directly searching the query text or by retrieving the same image and question from the public web. We curate candidates from three channels.

\paragraph{Public user-submitted arenas.}
We collect image-based user submissions from public vision-language model arenas, including VisionArena and WildVision \citep{chou2025visionarena,lu2024wildvision}. This source provides 284.4K images with diverse real-world scenes. We first apply an automated filter using \textsc{Claude-Opus-4.1} to remove images with limited visual information and cases that do not fit agentic problem settings. The filter also proposes a candidate task query that reflects the plausible action space. The prompt is shown in Appendix~\ref{apx:prompt_data_construction}. Human annotators then select high-quality candidates for downstream curation.

\paragraph{Annotator-captured real-life scenarios.}
We also include tasks collected by annotators from real daily situations, together with the original photos or screenshots that motivated the request. This channel naturally captures practical constraints, such as cluttered scenes, partial evidence, and ambiguous context, which are common in real deployments. We treat these instances as first-party user needs and keep their intent while ensuring the final task remains self-contained.

\paragraph{Private community forums.}
We also curate candidates from community help-seeking forums. We collect posts that include visually informative images and reflect realistic user goals. Since these posts often contain lengthy discussions and personal details, we rewrite each case into a standalone task while preserving the original intent and removing identifying information. We apply stricter screening to ensure clarity and consistency with our benchmark standards.

\section{Experimental Details}

\subsection{Tool Definition}
\label{sec:tool_definition_preface}
\textsc{AgentVista} is evaluated in a controlled tool environment with a compact set of commonly used tools for multimodal agent workflows. Models can invoke these tools appropriately within the \texttt{<tool\_call>...</tool\_call>} block during interaction. In detail, our tools are defined as follows.

\begin{tcolorbox}[breakable,
  colframe=bluegray,
  colback=bluegray!8,
  coltitle=white,
  title=Tool: Web Search]
\textbf{Description:} Search the web for information online. Use when you need to find information, facts, or current events. Returns web search results with titles, URLs, and text snippets. You only have limited search times, so please use it wisely. \\[5pt]
\textbf{Parameters:} 
\vspace{-5pt}
\begin{itemize}[leftmargin=*, itemsep=0pt]
    \item \textbf{query}: Search query string (required)
    \item \textbf{max\_results}: Maximum number of results to return (default: 10)
\end{itemize}
\end{tcolorbox}

\vspace{10pt}

\begin{tcolorbox}[breakable,
  colframe=bluegray,
  colback=bluegray!8,
  coltitle=white,
  title=Tool: Image Search]
\textbf{Description:} Search for related images using text query or reverse image search.\\
- For text-to-image search: specify search\_type=``text" and provide a query.\\
- For reverse image search: specify search\_type=``reverse" and provide an image\_url.\\
Returns images with URLs and descriptions. \\[5pt]
\textbf{Parameters:} 
\vspace{-5pt}
\begin{itemize}[leftmargin=*, itemsep=0pt]
    \item \textbf{search\_type}: Type of search ('text' or 'reverse', default: 'text')
    \item \textbf{query}: Search query string (required for 'text' mode)
    \item \textbf{image\_url}: Image filename, local reference, or URL (required for 'reverse' mode)
    \item \textbf{max\_results}: Maximum number of image results to return (default: 10)
\end{itemize}
\end{tcolorbox}

\vspace{10pt}

\begin{tcolorbox}[breakable,
  colframe=bluegray,
  colback=bluegray!8,
  coltitle=white,
  title=Tool: Visit]
\textbf{Description:} Visit a webpage and extract its main content. Use when you have a specific URL to visit (often after getting a URL from web search or image search). Extracts and returns the main textual content of the webpage. \\[5pt]
\textbf{Parameters:} 
\vspace{-5pt}
\begin{itemize}[leftmargin=*, itemsep=0pt]
    \item \textbf{url}: Full URL of the webpage to visit (must start with `http://' or `https://')
    \item \textbf{goal}: What information you want to find on this page (helps focus the extraction)
\end{itemize}
\end{tcolorbox}

\vspace{10pt}

\begin{tcolorbox}[breakable,
  colframe=bluegray,
  colback=bluegray!8,
  coltitle=white,
  title=Tool: Code Interpreter]
\textbf{Description:} Executes Python code in a stateful Jupyter kernel and return results. \\

\vspace{-5pt}

\textbf{Capabilities:}
\begin{itemize}[leftmargin=*, itemsep=0pt]
    \item Use PIL, NumPy, or OpenCV to process or analyze images to improve understanding.
    \item Perform complex mathematical calculations or data manipulation.
    \item Code execution is persistent: variables and functions are stored and reusable in subsequent calls.
\end{itemize}

\textbf{Image Variables (Pre-loaded as PIL Image objects):}
\begin{itemize}[leftmargin=*, itemsep=0pt]
    \item \textbf{original\_image / original\_image\_N (N=1,2,3,...):} The original input images are already loaded as PIL Image objects.
    \item \textbf{tool\_image\_N (N=1,2,3,...):} Images generated by your previous code calls.
    \item \textbf{observation\_N (N=1,2,3,...):} Images generated by zoom tool operations.
\end{itemize}

\textbf{Visualization \& Output:}
\begin{itemize}[leftmargin=*, itemsep=0pt]
    \item To view generated images, return the object, use \texttt{plt.show()}, or save it (e.g., \texttt{image.save('out.png')}).
    \item DO NOT use \texttt{image.show()}.
    \item Displayed images will be available as \texttt{"tool\_image\_N"} in the next turn.
\end{itemize}

\textbf{Pre-installed packages:}
\begin{itemize}[leftmargin=*, itemsep=0pt]
    \item PIL, NumPy, OpenCV, Matplotlib, SciPy, Scikit-learn, Scikit-image, Pandas, SymPy
\end{itemize}

\textbf{Parameters:} 
\vspace{-5pt}
\begin{itemize}[leftmargin=*, itemsep=0pt]
    \item \textbf{code}: Python code to execute in the Jupyter kernel (required)
\end{itemize}

\end{tcolorbox}

\subsection{Analysis of open-source model results.}
\label{apx:open_source}
\begin{table}[htbp]
  \caption{Results of representative open-source models on \textsc{AgentVista} by category. Domain abbreviations:
  \textbf{\textsc{Comm.}} (Commerce), \textbf{\textsc{Geog.}} (Geography), \textbf{\textsc{Ent.}} (Entertainment),
  \textbf{\textsc{Tech.}} (Technology), \textbf{\textsc{Soc.}} (Society), \textbf{\textsc{Acad.}} (Academics),
  and \textbf{\textsc{Cult.}} (Culture).
  The best-performing model in each category is \textbf{in-bold}, and the second best is \underline{underlined}. All values are accuracies in \%.}
  \small
  \label{tab:opensource_results}
  \centering
  \begin{tabular}{l|ccccccc|c}
    \toprule
    \textbf{\textsc{Model}} 
    & \textbf{\textsc{Comm.}} & \textbf{\textsc{Geog.}} & \textbf{\textsc{Ent.}} & \textbf{\textsc{Tech.}} & \textbf{\textsc{Soc.}} & \textbf{\textsc{Acad.}} & \textbf{\textsc{Cult.}}
    & \textbf{\textsc{Overall}} \\
    \midrule
    \textsc{Qwen3-VL-235B}  & \underline{7.14} & \underline{7.69} & \textbf{7.69} & \textbf{26.47} & \underline{16.00} & \textbf{20.00} & \underline{13.33} & \textbf{12.92} \\
    \textsc{DeepEyes-v2-7B} & \textbf{9.52} & \textbf{10.26} & \underline{2.56} & 14.71 & \textbf{24.00} & \underline{6.67} & \textbf{20.00} & \underline{11.48} \\
    \textsc{WebWatcher-32B} & 0.00 & \textbf{10.26} & 0.00 & \underline{23.53} & \textbf{24.00} & \textbf{20.00} & 0.00 & 10.05 \\
    \bottomrule
  \end{tabular}
\end{table}

Table~\ref{tab:opensource_results} reports results for three representative open-source multimodal models. In particular, \textsc{DeepEyes-v2-7B}~\cite{hong2025deepeyesv2} and \textsc{WebWatcher-32B}~\cite{geng2025webwatcher} are tool-using open-source agents that can interact with external tools to support multi-step problem solving, while \textsc{Qwen3-VL-235B} serves as a strong open-source multimodal backbone. Overall, these open-source baselines remain far from solving \textsc{AgentVista}, i.e., their overall accuracy ranges from 10.05\% to 12.92\%, substantially lower than the best-performing model \textsc{Gemini-3-Pro} at 27.3\%. This gap further reflects the ultra-challenging nature of \textsc{AgentVista} and highlights the large room for improving open-source multimodal agents.

\subsection{Prompts}
\subsubsection{Prompts for Data Construction}
\label{apx:prompt_data_construction}

\begin{tcolorbox}[breakable,
  colframe=sage,
  colback=sage!10,
  coltitle=white, title=Agentic Benchmark Task Filtering Criteria]
\textbf{Objective:} You are tasked with evaluating whether images are suitable for designing agent tasks with VERIFIABLE, UNIQUE ANSWERS, which require MULTIPLE SEARCHES and COMPLEX MULTI-HOP REASONING.\\

\textbf{Core Requirements (All must be satisfied):} 
\begin{itemize}[leftmargin=*, itemsep=1pt, topsep=1pt]
    \item \textbf{Complex Visual Content:} Rich, detailed visual elements (e.g., product catalogs, maps, menus, data tables, schedules) that support real-world task formulation.
    \item \textbf{Multiple Searches Required:} At least 2-3 distinct search operations are needed to gather diverse external information for solving the task.
    \item \textbf{Complex Multi-hop Reasoning:} The task must involve at least 4-5 reasoning steps that build on one another to arrive at the final solution.
    \item \textbf{Tool Synergy:} Tasks must leverage [Multiple Search/Browser operations + Complex Code execution] for an optimal solution.
    \item \textbf{Real-world Scenarios:} Tasks should be based on real-world scenarios, such as travel planning, shopping decisions, event scheduling, or data analysis.
    \item \textbf{Verifiable Unique Answer:} The task must lead to a verifiable and unique answer, which can be one of the following:
    \begin{itemize}[leftmargin=*, itemsep=1pt, topsep=1pt]
        \item A number (e.g., ``42", ``3.14", ``125", ``15km")
        \item A short string (e.g., ``Brittany", ``October 2, 2025", ``Paris")
        \item A date/time (e.g., ``2025-10-02", ``14:30", ``March 15")
        \item A location name (e.g., ``Central Park", ``Tokyo Station")
        \item A product/item name (e.g., ``iPhone 15", ``Margherita Pizza")
    \end{itemize}
\end{itemize}

\textbf{Must Reject:} 
\begin{itemize}[leftmargin=*, itemsep=1pt, topsep=1pt]
    \item \textbf{OCR Tasks:} Pure text recognition, translation, or transcription.
    \item \textbf{Simple Images:} Anime characters, simple landscapes, basic objects, or cartoons.
    \item \textbf{Academic Tasks:} Tasks involving papers, formulas, or professional charts.
    \item \textbf{Direct Q\&A:} Questions that can be answered directly without the need for external tools.
    \item \textbf{Too Simple:}
    \begin{itemize}[leftmargin=*, itemsep=1pt, topsep=1pt]
        \item Less than 3 reasoning steps.
        \item Single-hop reasoning (e.g., ``Find the price of X").
        \item Simple calculations that lack complex logic.
    \end{itemize}
    \item \textbf{Non-verifiable Answers:}
    \begin{itemize}[leftmargin=*, itemsep=1pt, topsep=1pt]
        \item Plans or itineraries (e.g., ``Design a travel plan").
        \item Reports or summaries (e.g., ``Write a shopping report").
        \item Recommendations or suggestions (e.g., ``Recommend 3 hotels").
        \item Long explanations or descriptions.
        \item Multiple answers or lists (unless asking for a count).
        \item Subjective opinions.
    \end{itemize}
\end{itemize}
\end{tcolorbox}

\subsubsection{The Prompt for Evaluation}

\begin{tcolorbox}[breakable,
  colframe=sage,
  colback=sage!10,
  coltitle=white,
  title=Multimodal Agent Prompt]
You are a visual reasoning agent. Your goal is to answer questions about images.\\

\textbf{AVAILABLE TOOLS:}
\begin{itemize}[leftmargin=*, itemsep=1pt, topsep=1pt]
    \item \textbf{web\_search:} Search the web for information, facts, or current events.
    \item \textbf{image\_search:} Search for related images using text query or reverse image search.
    \item \textbf{visit:} Visit a webpage and extract its main content.
    \item \textbf{code\_interpreter:} Execute Python code for image processing, analysis, and calculations.
\end{itemize}
\vspace{5pt}
\textbf{INSTRUCTIONS:}
\begin{enumerate}[leftmargin=*, itemsep=1pt, topsep=1pt]
    \item \textbf{Analyze:} Carefully observe the image and the user's question.
    \item \textbf{Think:} Explain your step-by-step reasoning process.
    \item \textbf{Use Tools:} Call the appropriate tool to gather information and help answer the question.
    \item \textbf{Iterate as needed:} Continue reasoning and using tools in next turns until you are confident in your findings.
    \item \textbf{Answer:} Once confident, provide the final answer inside \texttt{<answer>...</answer>} tags!
\end{enumerate}
\vspace{5pt}
\textbf{IMPORTANT:}
\begin{itemize}[leftmargin=*, itemsep=1pt, topsep=1pt]
    \item Always explain your detailed reasoning process before using any tool.
    \item You can ONLY call one tool at a time! Do not call multiple tools in one turn!
    \item You MUST provide your final answer using complete \texttt{<answer>...</answer>} tags!
\end{itemize}

\end{tcolorbox}

\section{Error type definitions.}
\label{apx:error}
In Section~\ref{sec:error_analysis}, we report the error distributions of representative models on \textsc{AgentVista}. Here we define the error types used in our taxonomy.
\vspace{-5pt}
\paragraph{Tool execution failure.}
This category captures cases where the agent follows a plan, but fails due to issues in tool interaction. Typical examples include empty tool outputs, invalid requests, and failures to open or parse retrieved content. These errors suggest that robust tool use and self-checking are important for completing long-horizon workflows.

\paragraph{Visual misidentification.}
This category includes errors caused by incorrect visual understanding, such as reading the wrong text on a label, confusing similar components, missing a small indicator, or miscounting objects. Because visual evidence often determines what to search for and how to apply constraints, a single perception mistake can cause later steps to follow an incorrect direction.

\paragraph{Knowledge hallucination.}
This category refers to cases where the agent outputs facts that are not supported by the provided images or retrieved sources. Common patterns include inventing details that look plausible, relying on generic rules of thumb, or asserting standards that do not match the evidence in the current instance. These failures indicate insufficient grounding in the multimodal context.

\paragraph{Calculation error.}
This category covers mistakes in arithmetic or multi-step aggregation, such as wrong unit conversions, incorrect date computations, or errors when combining multiple retrieved values. These cases often arise after several steps, when the agent must keep intermediate numbers consistent while continuing to use tools.

\paragraph{Instruction misinterpretation.}
This category includes failures to follow the user request or constraints, such as ignoring a time window, missing a required format, applying the wrong condition, or answering a related but different question. Even when perception and retrieval are correct, misunderstanding the intent can still lead to an incorrect final answer.

\paragraph{Others.}
This category groups remaining failures that do not fit the above types or that involve multiple types without a clear primary cause. Examples include incomplete final answers, premature termination, inconsistent outputs across steps, or cases where the model produces an answer that cannot be checked against the required format. We use this bucket to keep the taxonomy simple while still accounting for long-tail error patterns.

\section{Case Study}
In this section, we present representative trajectories to illustrate both successful and failed behaviors on \textsc{AgentVista}. We first show a good-case example that demonstrates effective long-horizon, interleaved tool use. We then provide one bad-case example for each error type, highlighting how different failure modes arise and how they derail the overall workflow.

\label{apx:case_study}
\subsection{Good Case Examples}

\paragraph{Traj \#1: Sneaker Authentication.} This task involved verifying the authenticity of luxury sneakers based on visual evidence. Through a sequence of seven tool invocations, the model conducted a systematic examination of specific features. It utilized Image Search to contrast tongue and size tags with authentic references, identifying an anomalous "A8513" sticker. Subsequent validation via Web Search confirmed this as a counterfeit indicator, leading to the correct classification.
\paragraph{Traj \#2: Strongest German Beer Analysis.} Identifying the strongest beer required distinguishing specific brands within a cluttered image. The model synergized the Code Interpreter for visual refinement with Web Search for factual retrieval. This approach enabled the precise filtering of lower-alcohol options, resulting in the accurate identification of a tie between Steam Brew German Red and Perlenbacher Strong.

\begin{tcolorbox}[breakable,
  colframe=dustyrose,
  colback=white,
  coltitle=white,
  title={Traj \#1: Sneaker Authentication (\textsc{Gemini-3-Pro}; success; 7 tool calls)},
  fonttitle=\bfseries, before skip=1mm]
\small

\textbf{Task.} Acting as a luxury sneaker authenticator, search for authentic craftsmanship images for comparison. Decide whether the shoes are \textbf{Real} or \textbf{Fake}, and give at least two visual reasons.

\par\vspace{1mm}
\textit{Input images.}
\begin{center}
    \includegraphics[height=3.6cm]{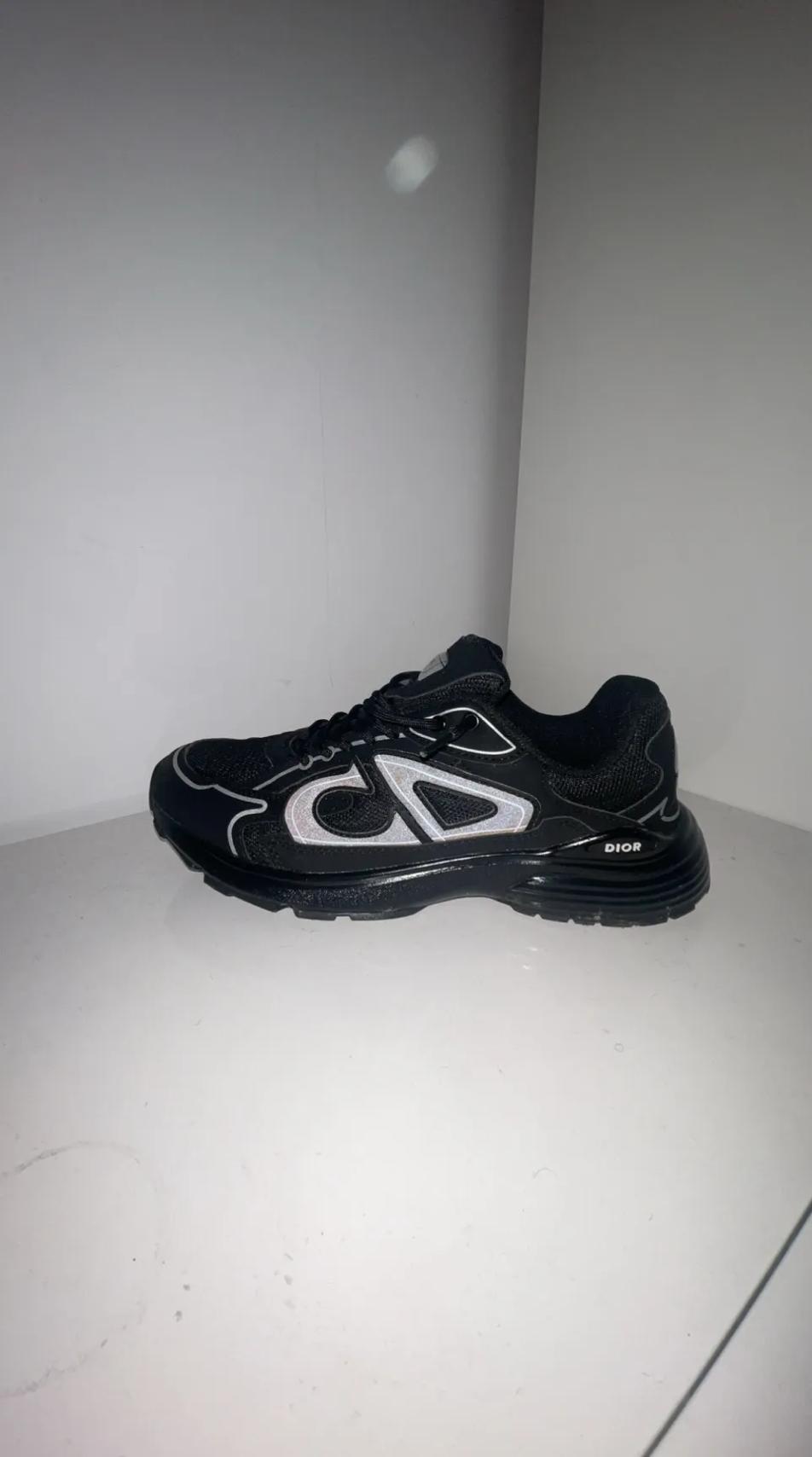}\hspace{1mm}
    \includegraphics[height=3.6cm]{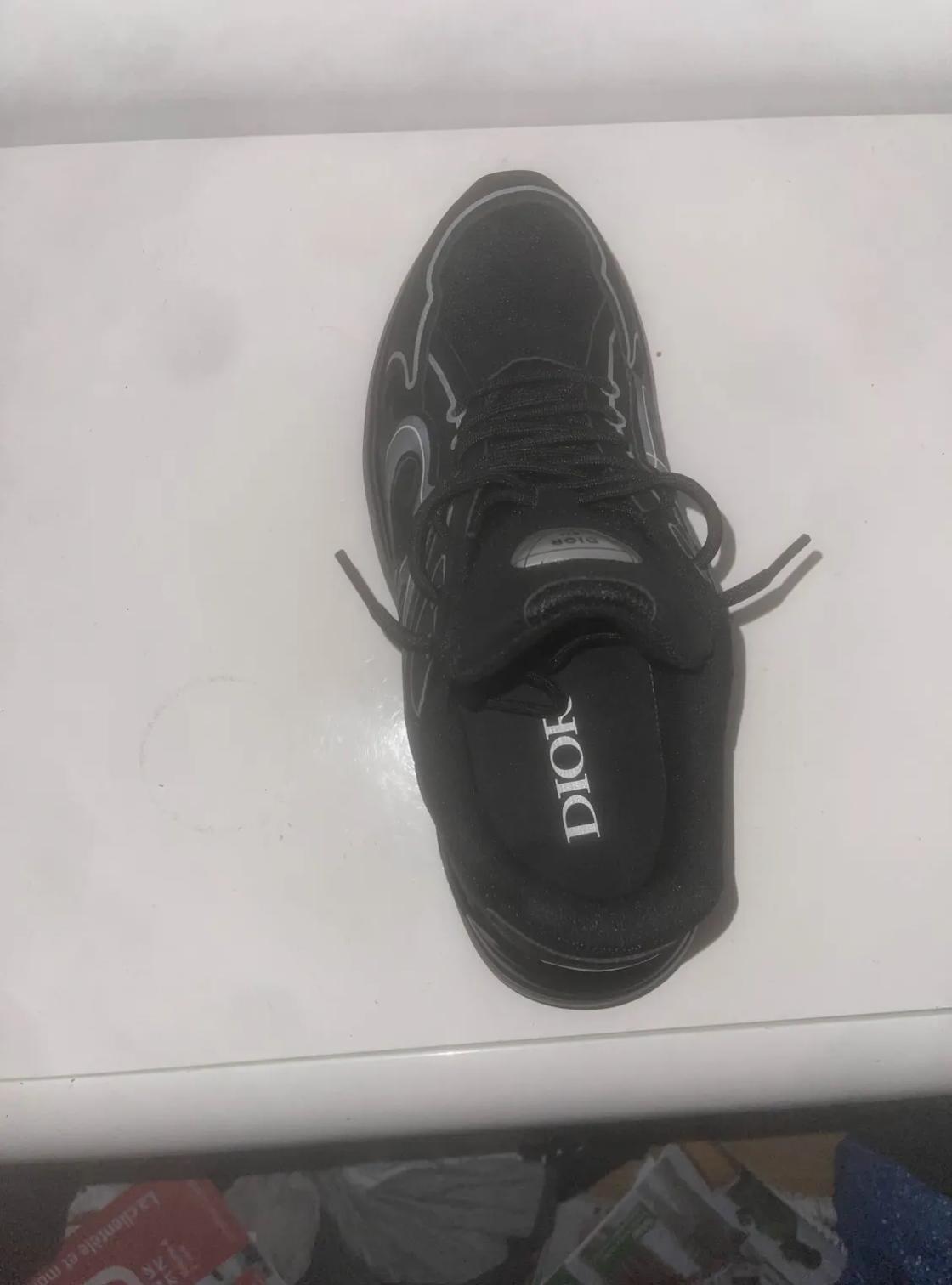}\hspace{1mm}
    \includegraphics[height=3.6cm]{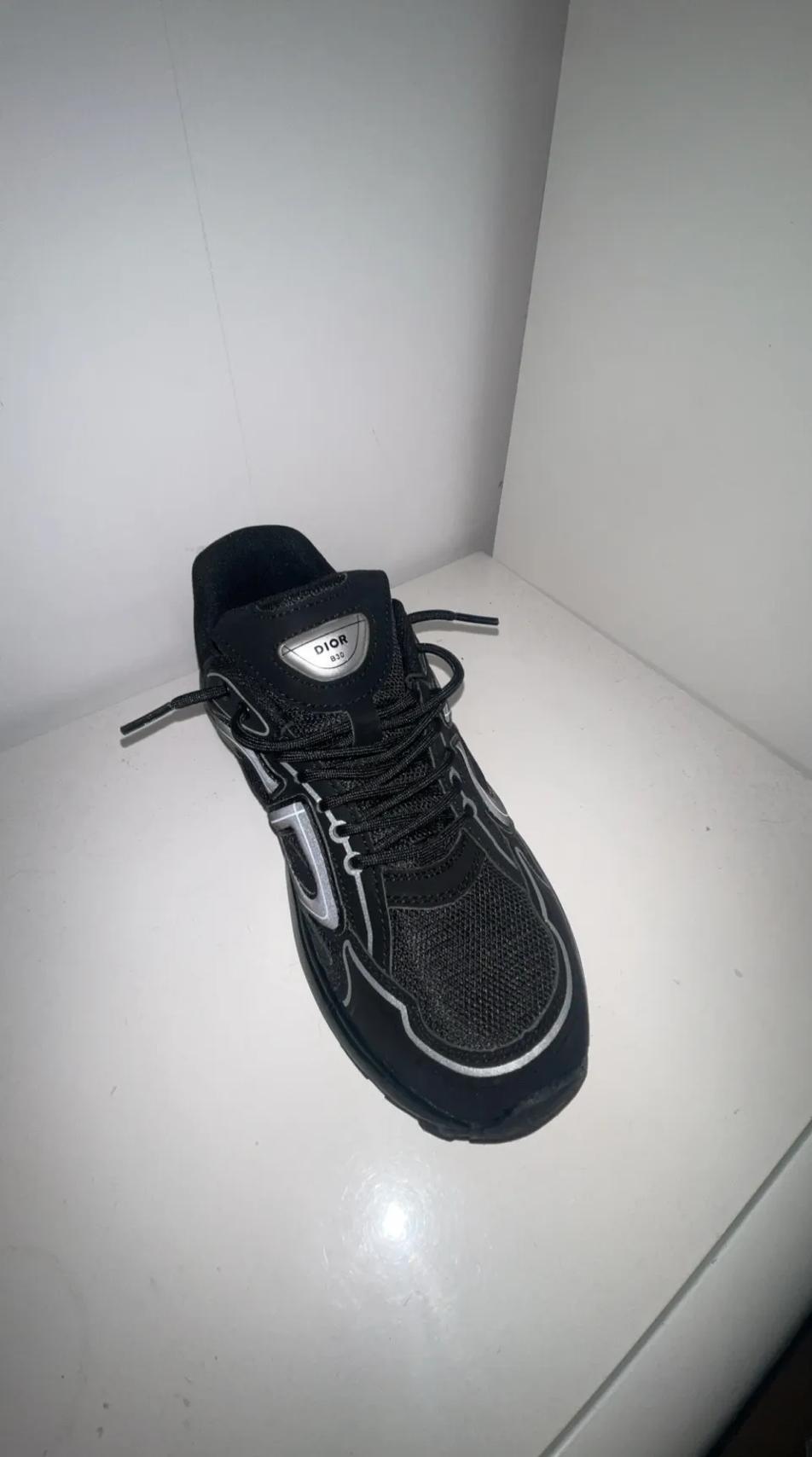}\hspace{1mm}
    \includegraphics[height=3.6cm]{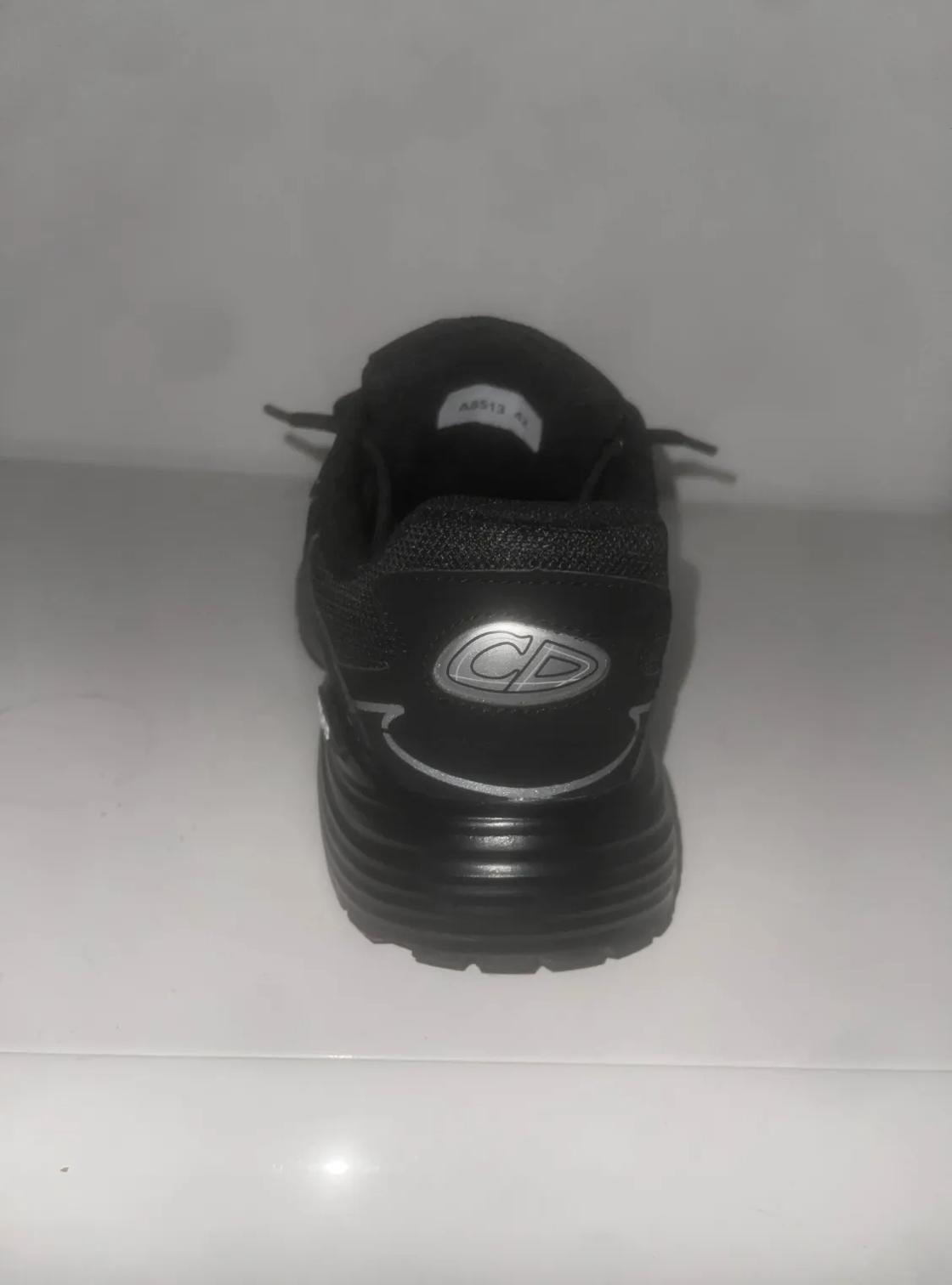}
\end{center}

\par\vspace{2mm}
\textbf{Ground truth.} Counterfeit. Key cues include Widow's Peak Error, Poor Tongue Font \& Stitching, Misaligned 3M Stripes.
\par\vspace{2mm}\hrule

\vspace{2mm}
\textbf{Tool trajectory (condensed).}

\textbf{Step 1.} \textit{web search} \\
Query: \texttt{Dior B30 black sneaker authentic vs fake guide} \\
\textbf{Evidence.} Retrieved legit-check guides that highlight tongue tag font, stitching, and interior label formats as key checks.

\vspace{1mm}
\textbf{Step 2.} \textit{image search} \\
Query: \texttt{authentic Dior B30 black sneaker tongue tag} \\
\textbf{Evidence.} Collected reference images for tongue tag layout and font.
\begin{center}
\includegraphics[height=3.1cm]{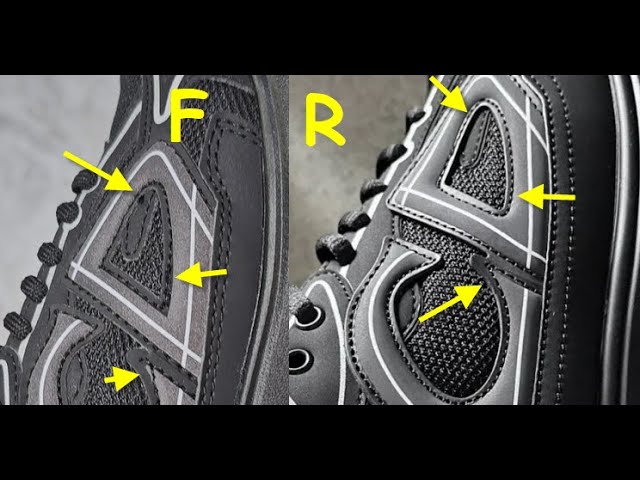}\hspace{1mm}
\includegraphics[height=3.1cm]{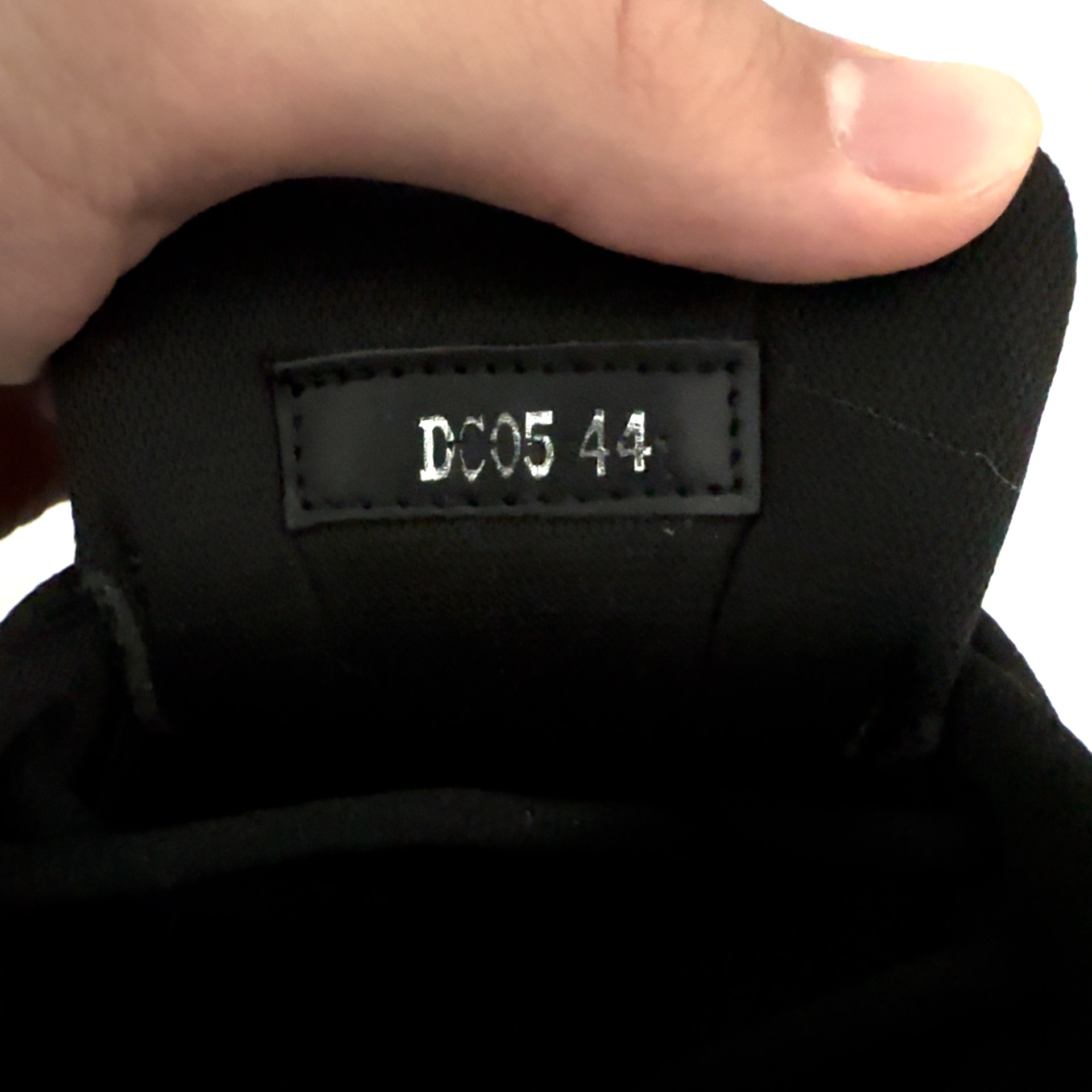}
\end{center}

\vspace{1mm}
\textbf{Step 3.} \textit{image search} \\
Query: \texttt{authentic Dior B30 sneaker inside size tag label} \\
\textbf{Evidence.} Collected reference images for inside label formats and placement.
\begin{center}
\includegraphics[height=3.1cm]{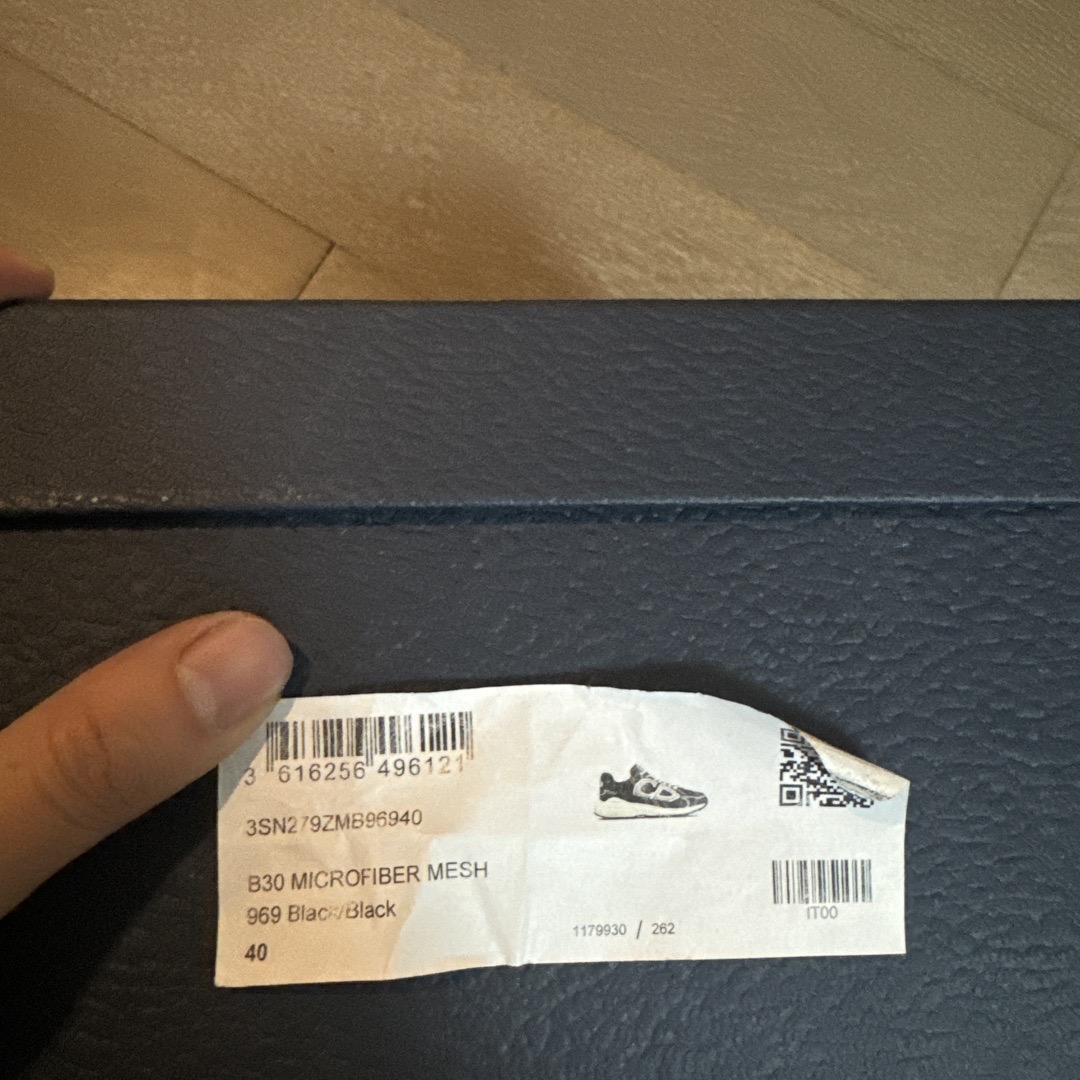}\hspace{1mm}
\includegraphics[height=3.1cm]{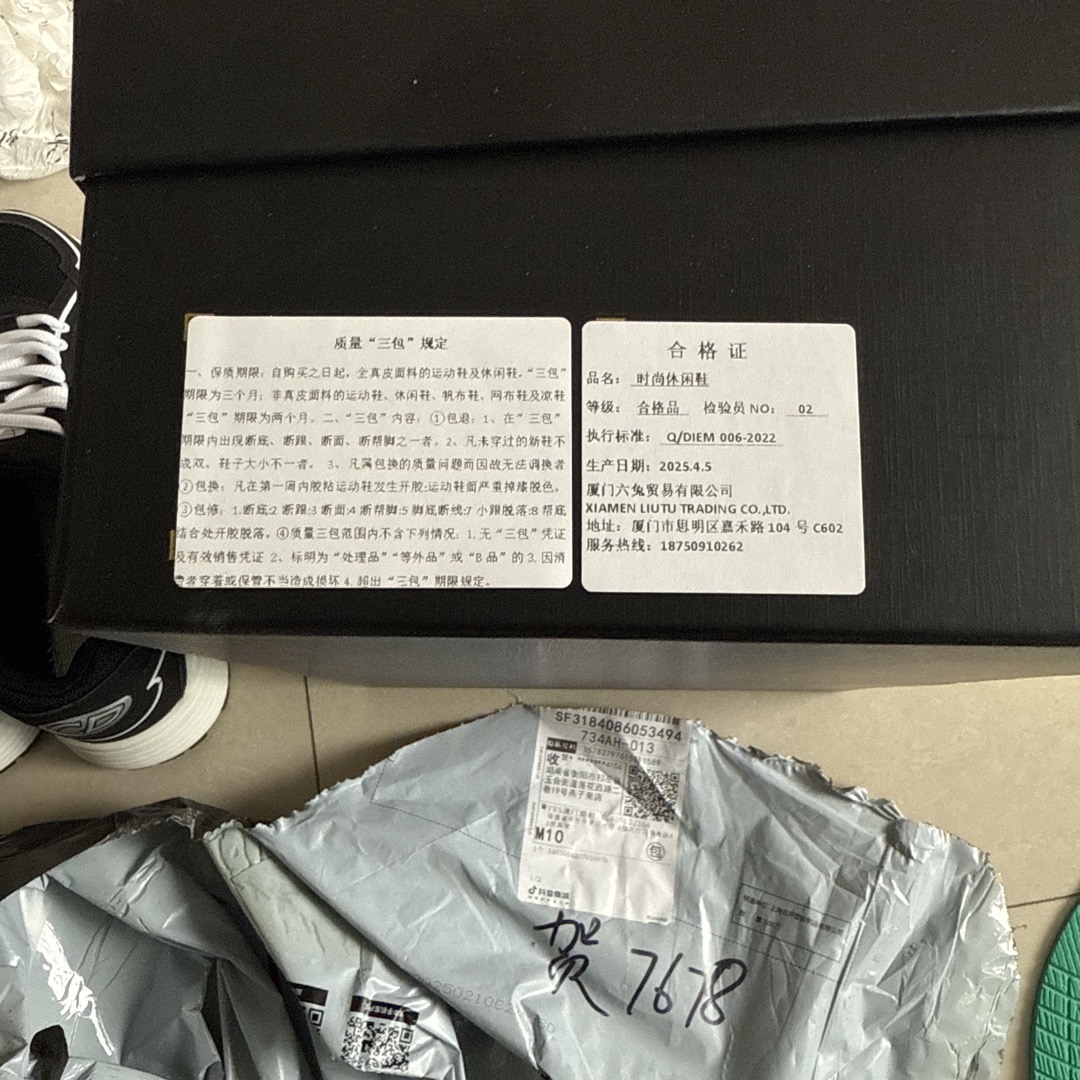}
\end{center}

\vspace{1mm}
\textbf{Step 4.} \textit{web search} \\
Query: \texttt{Dior B30 authentic inner size code format} \\
\textbf{Evidence.} Guides report authentic style identifiers follow a structured style-ID pattern, rather than generic sticker-like codes.

\vspace{1mm}
\textbf{Step 5.} \textit{image search} \\
Query: \texttt{fake Dior B30 size sticker A8513} \\
\textbf{Evidence.} Collected examples where generic interior stickers are associated with replicas.
\begin{center}
\includegraphics[height=3.1cm]{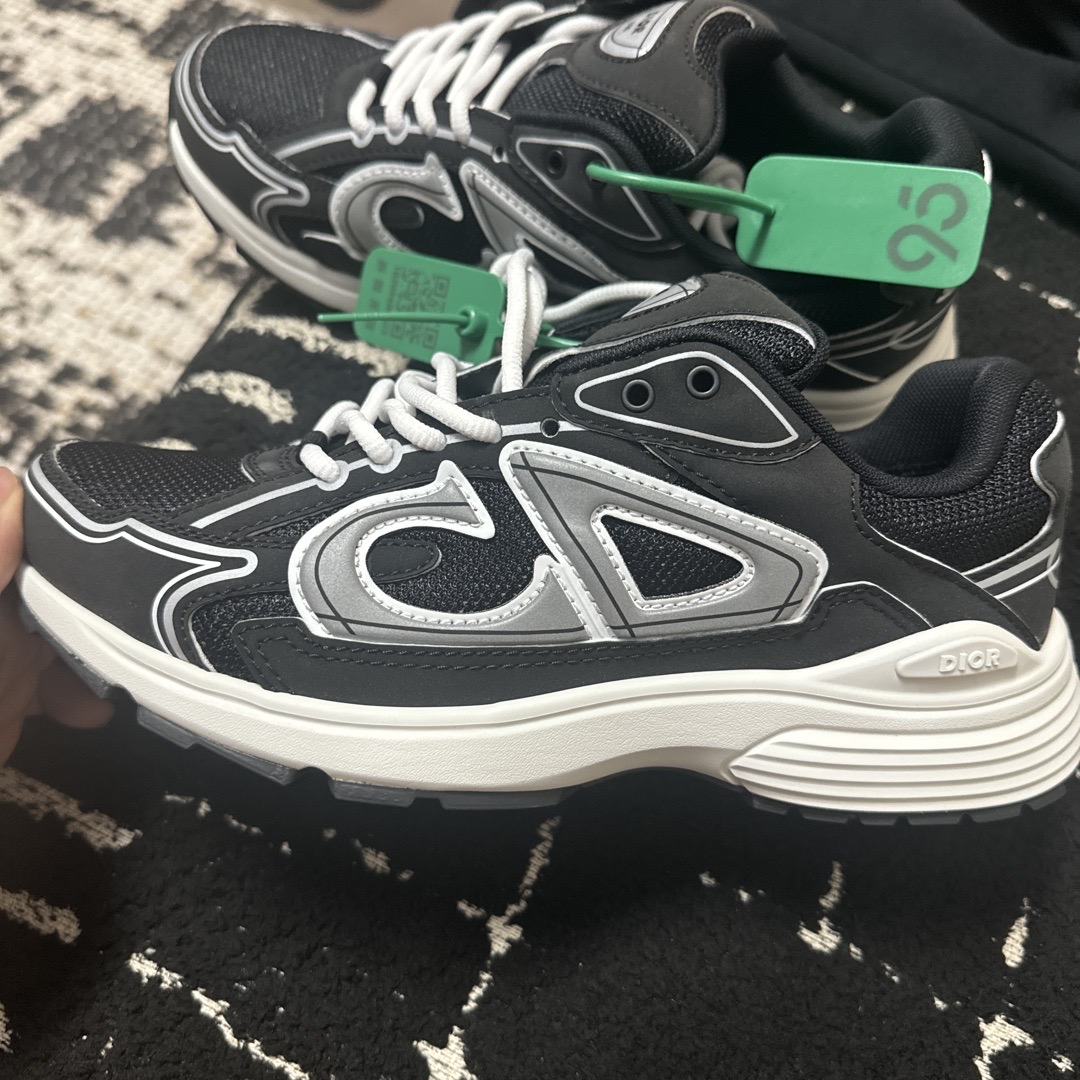}\hspace{1mm}
\includegraphics[height=3.1cm]{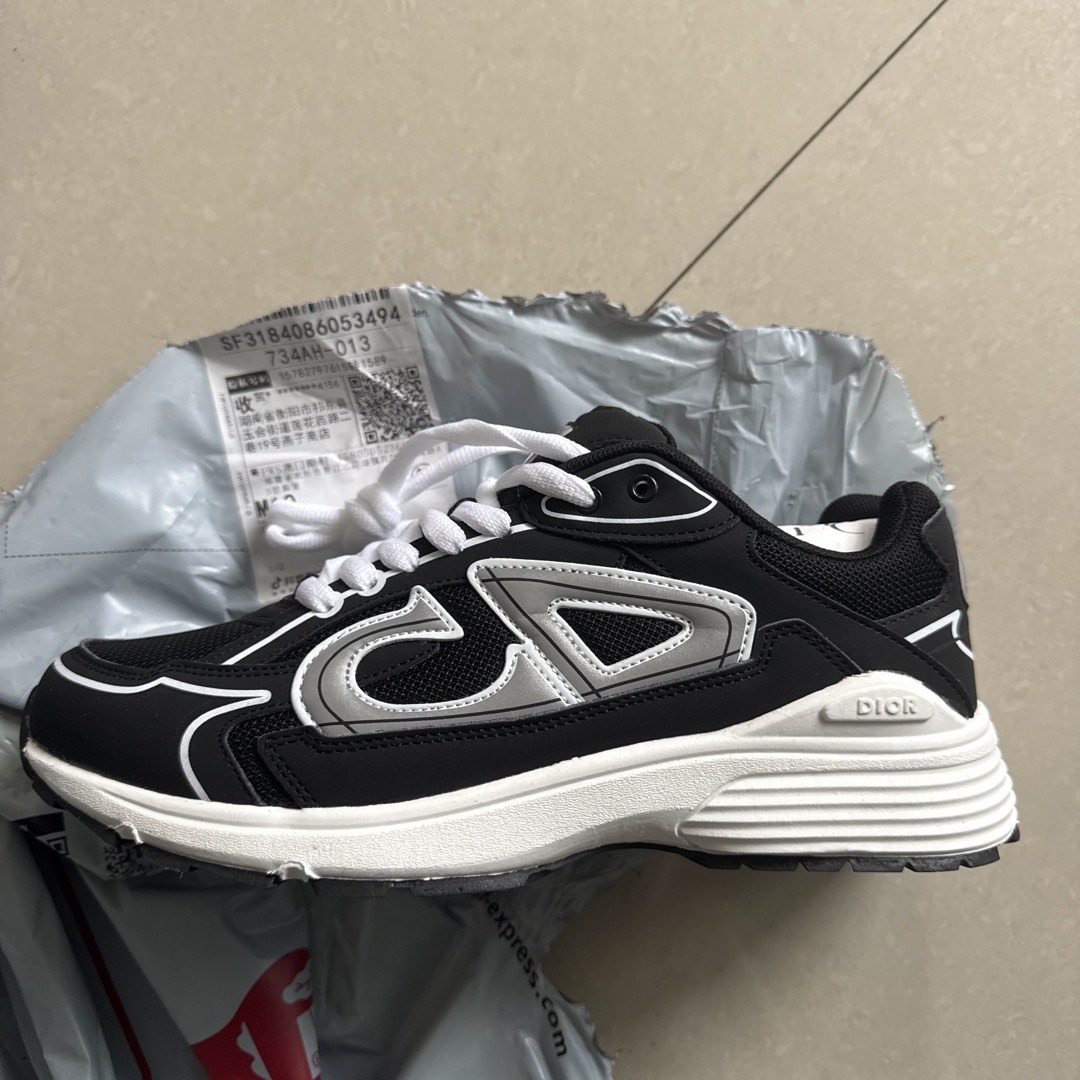}
\end{center}

\vspace{5pt}
\par\hrule\vspace{2mm}

\textbf{Final decision.} \textbf{Fake.} \\
\textbf{Reasons.}
\vspace{-5pt}
\begin{itemize}[leftmargin=*]
\setlength{\itemsep}{0pt}
\item The interior sticker code format is inconsistent with authentic labeling patterns shown in retrieved references.
\item The tongue tag font and stitching do not match the authentic reference images. \Yes
\end{itemize}

\end{tcolorbox}

\begin{tcolorbox}[breakable,
  colframe=dustyrose,
  colback=white,
  coltitle=white,
  title={Traj \#2: Strongest German Beer Analysis (\textsc{Gemini-3-Pro}; success; 7 tool calls)},
  fonttitle=\bfseries]
\small

\textbf{Task.} From the beers shown in the photo, consider only German-brewed beers with ABV $>$ 5\%. Which \textbf{brand} has the highest \textbf{total alcohol per can}, accounting for both ABV and can volume?

\par\vspace{1mm}
\textit{Input image.}
\begin{center}
    \includegraphics[height=3.6cm]{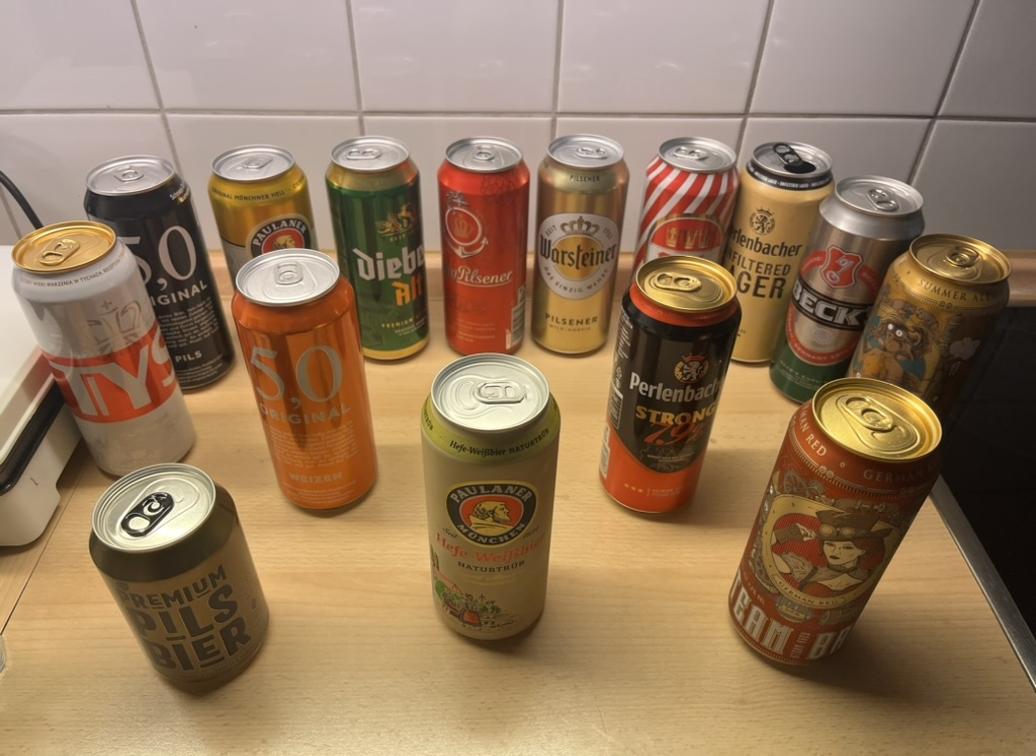}
\end{center}

\par\vspace{2mm}
\textbf{Ground truth.} \textbf{Steam Brew German Red} and \textbf{Perlenbacher Strong} (tie).
\par\vspace{2mm}\hrule

\vspace{2mm}
\textbf{Tool trajectory (condensed).}

\textbf{Step 1.} \textit{code interpreter} \\
\textbf{Evidence.} Zoomed and cropped the photo to read brand names and the ABV/volume information on the cans.
\begin{center}
\includegraphics[height=3.0cm]{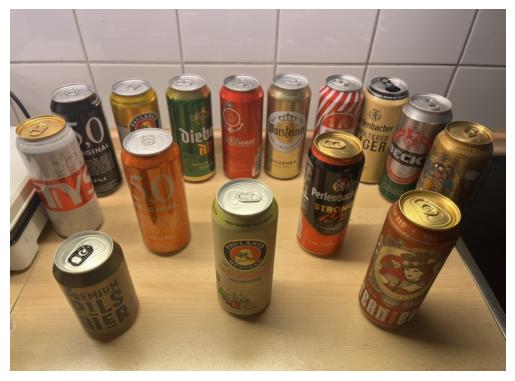}
\end{center}

\vspace{1mm}
\textbf{Step 2.} \textit{web search} \\
Query: \texttt{Perlenbacher Strong ABV \ \ Steam Brew German Red ABV} \\
\textbf{Evidence.} Retrieved product pages confirming both beers are \textbf{7.9\% ABV}.

\vspace{1mm}
\textbf{Step 3.} \textit{code interpreter} \\
\textbf{Evidence.} Cropped individual cans to verify they are German-brewed and to confirm can capacity (standard tall can).
\begin{center}
\includegraphics[height=3.0cm]{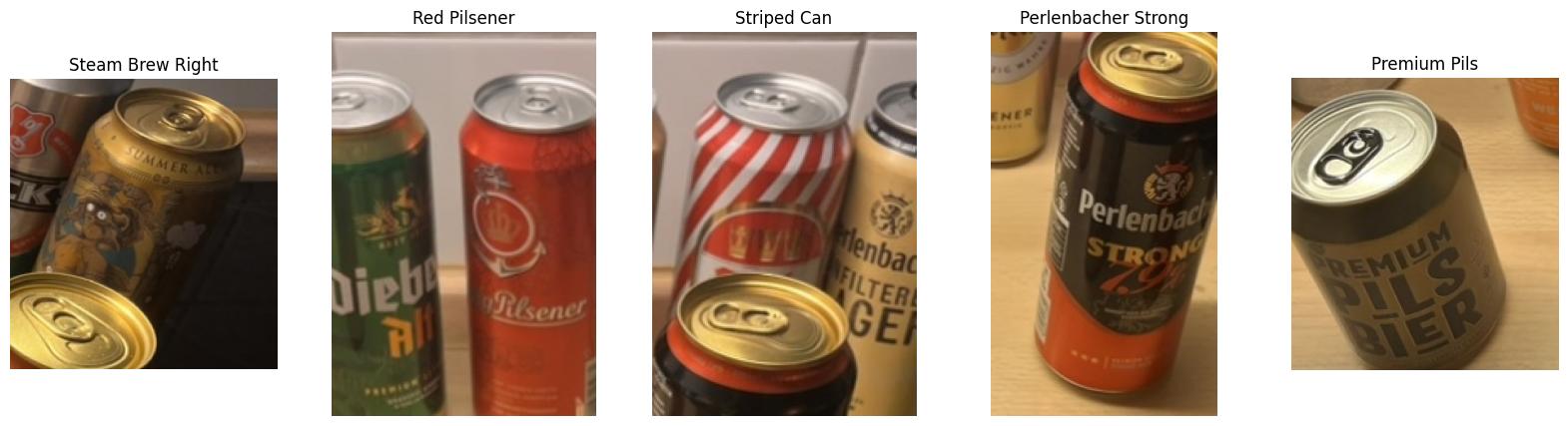}\hspace{1mm}
\includegraphics[height=3.0cm]{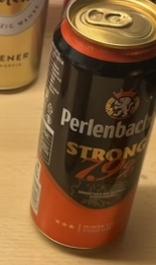}\hspace{1mm}
\includegraphics[height=3.0cm]{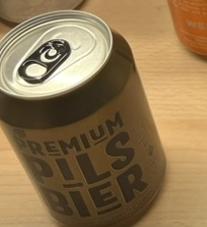}
\end{center}

\vspace{1mm}
\textbf{Step 4.} \textit{web search} \\
Query: \texttt{Steam Brew German Red volume 500ml} \\
\textbf{Evidence.} Confirmed the can size used for calculation is \textbf{500 ml}.

\vspace{1mm}
\textbf{Step 5.} \textit{image search} \\
\textbf{Evidence.} Attempted to identify other cans/variants; no higher-ABV German-brewed option was found among the eligible beers.

\vspace{1mm}
\textbf{Step 6.} \textit{web search} \\
Query: \texttt{Steam Brew German Red ABV exact \ \ Perlenbacher Strong ABV} \\
\textbf{Evidence.} Cross-checked ABV values and resolved remaining ambiguity.

\vspace{5pt}
\par\hrule\vspace{2mm}

\textbf{Final decision.} \textbf{Tie: Steam Brew German Red and Perlenbacher Strong.} \\
\textbf{Reason.} Both are German-brewed, exceed 5\% ABV, and have the same alcohol per can:
$0.079 \times 500\text{ ml} = 39.5\text{ ml}$. \Yes

\end{tcolorbox}


\clearpage
\subsection{Bad Case Examples}
\label{sec:bad_cases}

\paragraph{Traj \#3: Karst Jigsaw Puzzle. \textit{Tool execution failure}.}
\textbf{Task.} Reconstruct a $6{\times}6$ jigsaw puzzle from an input image and locate the missing piece position.
\textbf{Failure.} The model attempted to segment puzzle pieces with code-based image processing, but the segmentation failed and extracted only 24 segments instead of the expected 35. Without a complete set of pieces, the model could not form a valid grid and the reconstruction became infeasible.
\textbf{Classification Rationale.} The core issue is a breakdown in tool-based image processing, which blocks the workflow even though the high-level plan is reasonable.

\vspace{{-5pt}}
\paragraph{Traj \#4: Authors United Window Display. \textit{Visual misidentification}.}
\textbf{Task.} Identify the author shown in a window display from the provided image.
\textbf{Failure.} The visible author is Donna Tartt, but the model failed to identify her. Although it performed cropping, it still did not extract the correct visual cue and produced an incorrect identification.
\textbf{Classification Rationale.} The decisive evidence is in the image, and the failure comes from incorrect visual recognition rather than retrieval or reasoning.
\vspace{{-5pt}}
\paragraph{Traj \#5: Target Arena Identification. \textit{Visual misidentification}.}
\textbf{Task.} Identify the correct university basketball facility shown in the image.
\textbf{Failure.} The model misread an unclear floor logo and anchored on the wrong university, then reinforced the mistake using generic features such as roof trusses. It concluded the venue was St.\ Thomas AARC, while the correct answer is UNC.
\textbf{Classification Rationale.} The initial mistake is a wrong visual anchor, and later steps follow that incorrect anchor.
\vspace{{-5pt}}
\paragraph{Traj \#6: Pilea Root Diagnosis. \textit{Knowledge hallucination}.}
\textbf{Task.} Diagnose the hard mass on Pilea roots from the image.
\textbf{Failure.} The correct interpretation is calloused residue from root rot, but the model claimed it was a ``nursery plug'' or fungal material and described visual properties that are not supported by the image. The final diagnosis followed a made-up interpretation aligned with retrieval results rather than the provided evidence.
\textbf{Classification Rationale.} The model introduces unsupported facts and forces the image to fit a preconceived explanation.

\vspace{{-10pt}}
\paragraph{Traj \#7: Studio Swing Prop Design. \textit{Instruction misinterpretation}.}
\textbf{Task.} Design a stationary photo prop that visually looks like a suspended swing.
\textbf{Failure.} The model proposed a design where the seat is visibly supported by a horizontal bar, which removes the hanging illusion and violates the core constraint of the request.
\textbf{Classification Rationale.} The model fails to follow the key constraint and answers a different problem than the one asked.

\begin{tcolorbox}[breakable,
  colframe=dustyrose,
  colback=white,
  coltitle=white,
  title={Traj \#3: Karst Jigsaw Puzzle (\textsc{Gemini-3-Pro}; failed; 3 tool calls)},
  fonttitle=\bfseries]
\small

\textbf{Task.} The first image shows a $6{\times}6$ jigsaw puzzle (one piece missing) made from a karst-landform photo. The second image is a local reference landscape photo. Reconstruct the puzzle and report the coordinate of the missing piece. Coordinates are defined on the reconstructed grid: top-left is $(1,1)$, top-right is $(1,6)$, and bottom-right is $(6,6)$.

\par\vspace{1mm}
\textit{Input images.}
\begin{center}
  \includegraphics[height=3.4cm, keepaspectratio]{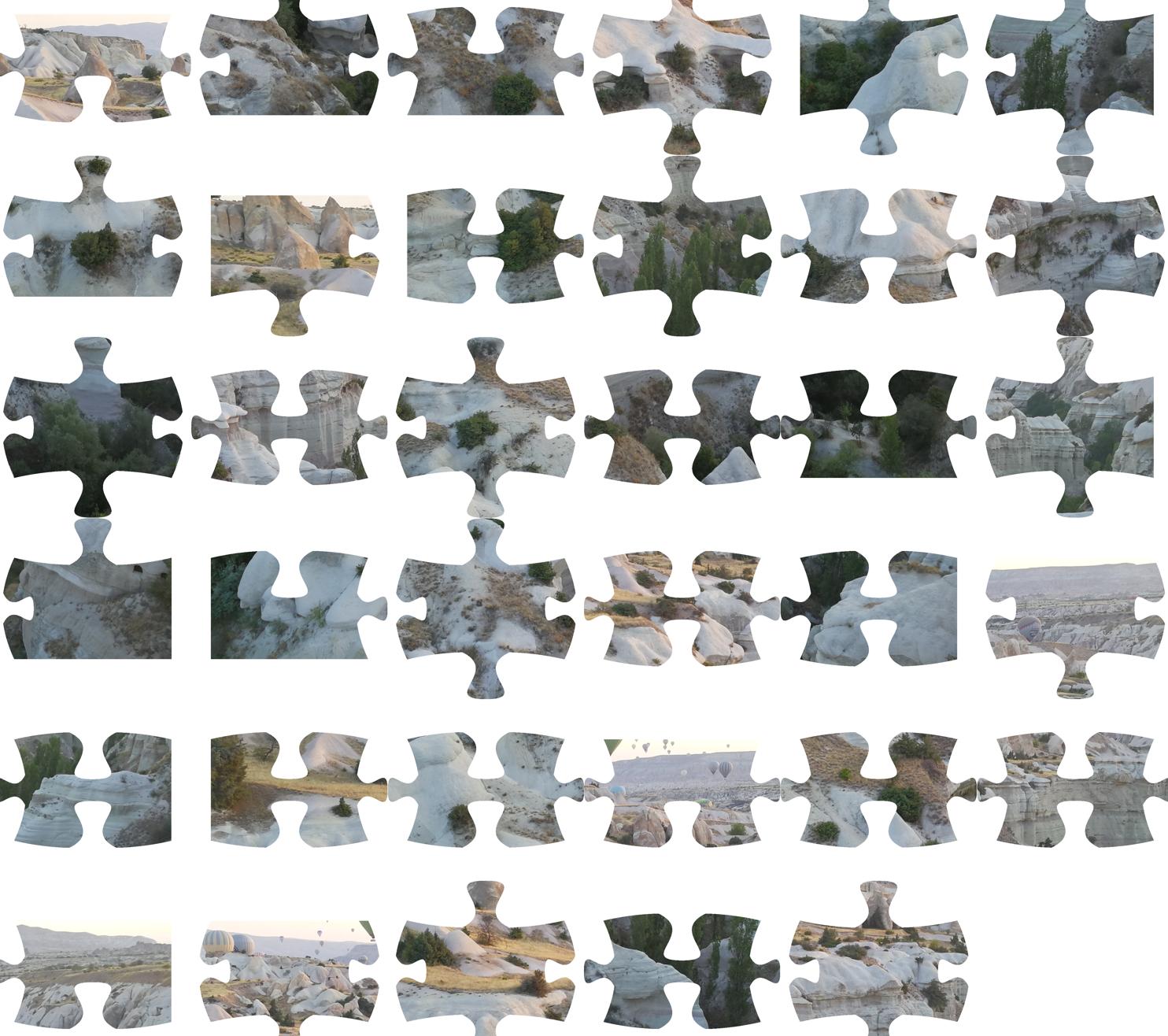}\hspace{1mm}
  \includegraphics[height=3.4cm, keepaspectratio]{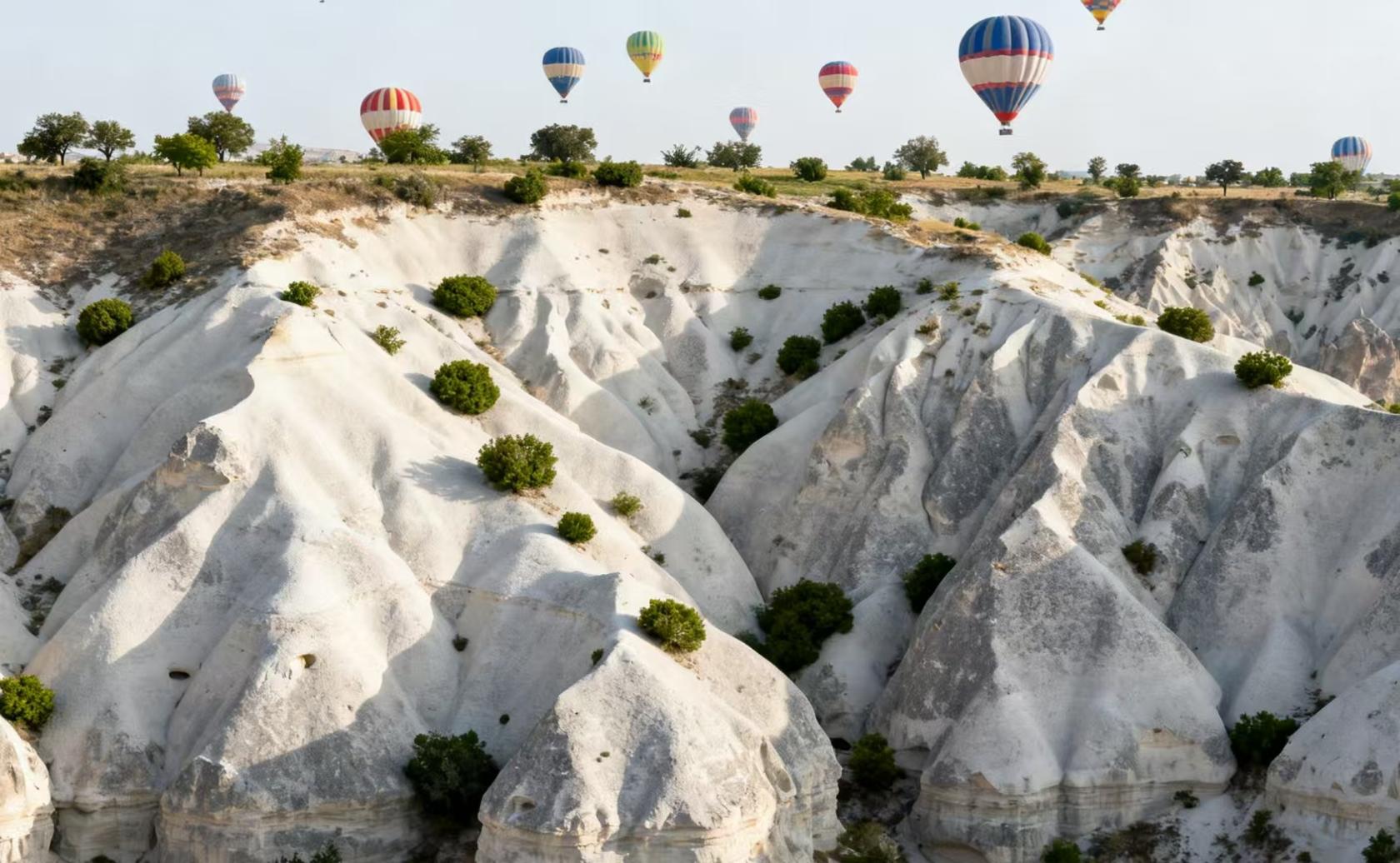}
\end{center}

\vspace{1mm}
\textbf{Ground truth.} $(3,1)$
\vspace{5pt}
\par\hrule\vspace{2mm}
\textbf{Tool trajectory (condensed).}

\textbf{Step 1.} \textit{code interpreter} \\
\textbf{Goal.} Segment the puzzle into individual pieces and assign each piece to a grid cell. \\
\textbf{Outcome.} The segmentation merges adjacent pieces and fails to recover all pieces.
\begin{center}
  \includegraphics[height=6.8cm, keepaspectratio]{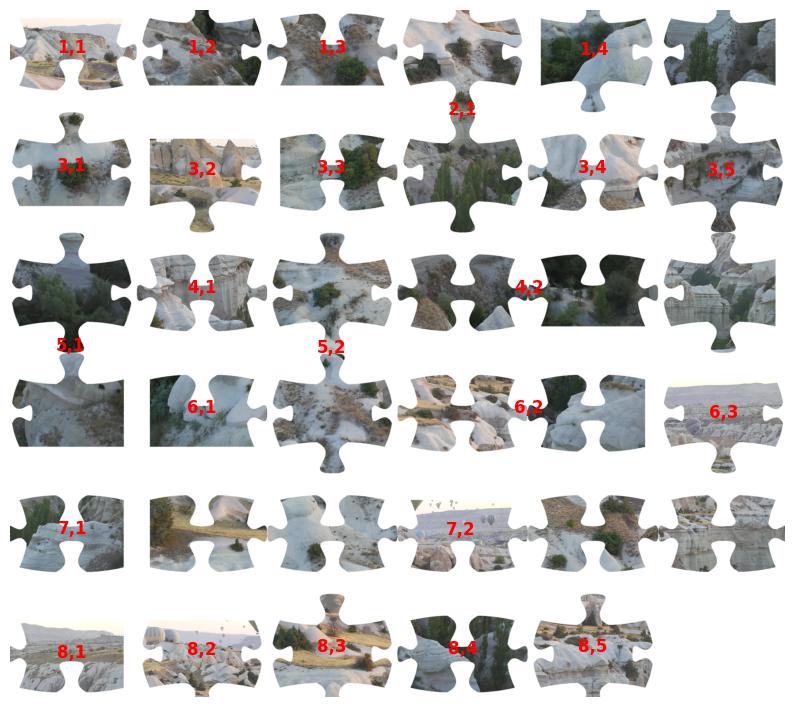}
\end{center}

\vspace{1mm}
\textbf{Step 2.} \textit{code interpreter} \\
\textbf{Goal.} Improve segmentation with morphological operations and contour filtering. \\
\textbf{Outcome.} Still under-segments the puzzle and extracts only \textbf{24} regions, instead of the expected \textbf{35} pieces (or 36 slots).
\begin{center}
  \includegraphics[height=6.8cm, keepaspectratio]{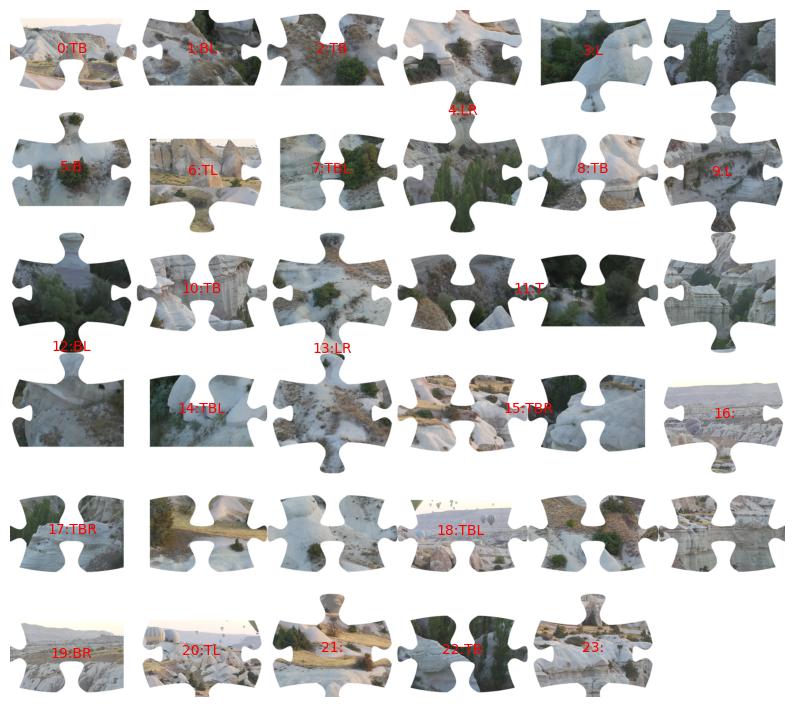}
\end{center}

\vspace{1mm}
\textbf{Step 3.} \textit{code interpreter} \\
\textbf{Goal.} Crop and visualize the extracted pieces for downstream matching. \\
\textbf{Outcome.} Visualization confirms missing segments; reconstruction cannot proceed.
\begin{center}
  \includegraphics[height=3.4cm, keepaspectratio]{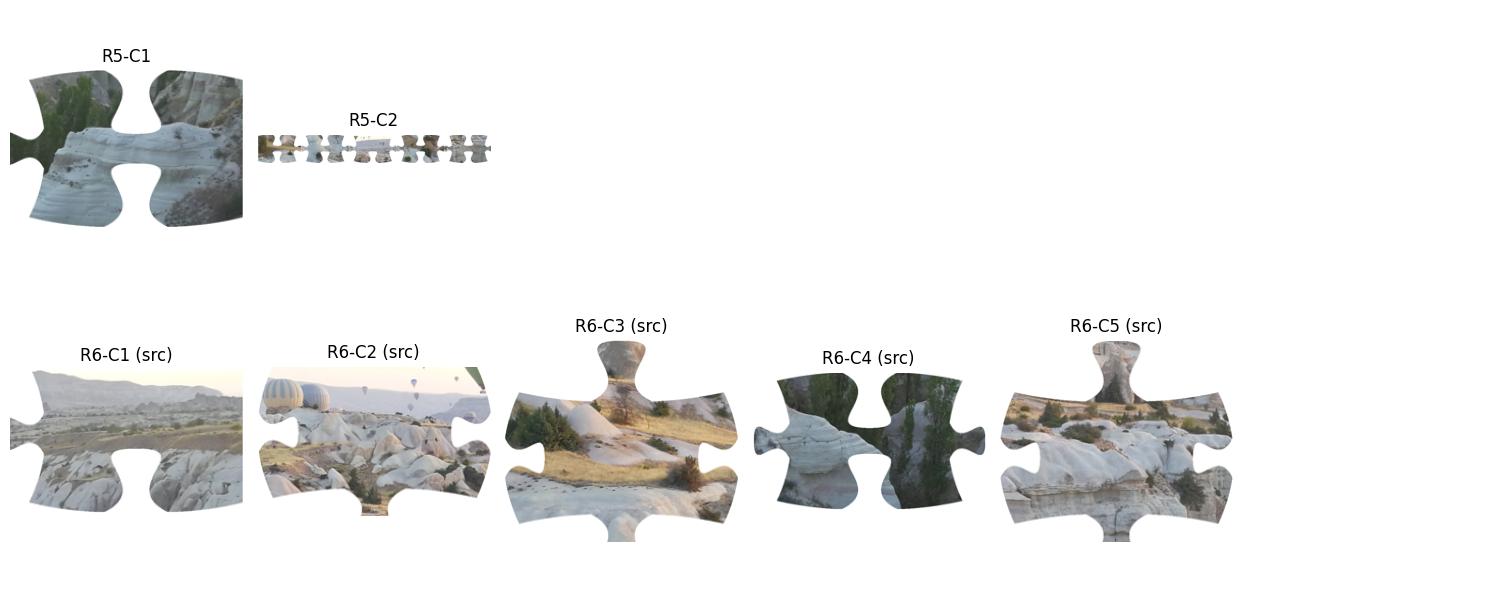}
\end{center}

\vspace{5pt}
\par\hrule\vspace{2mm}
\textbf{Failure point.} \textbf{Step 2 (segmentation).} The model cannot reliably separate touching pieces, so it fails to obtain a complete set of puzzle pieces. Without correct piece extraction, subsequent grid reconstruction and missing-cell identification are not feasible.

\vspace{5pt}
\par\hrule\vspace{2mm}
\textbf{Final outcome.} The model fails to reconstruct the $6{\times}6$ layout and cannot determine the missing coordinate. \No

\end{tcolorbox}

\begin{tcolorbox}[breakable,
  colframe=dustyrose,
  colback=white,
  coltitle=white,
  title={Traj \#4: Authors United Window Display (\textsc{Gemini-3-Pro}; failed; 13 tool calls)},
  fonttitle=\bfseries]
\small

\textbf{Task.} This photo shows a bookstore window display supporting ``Authors United,'' featuring photos of authors who have appeared at the bookstore. Among the authors whose photos are clearly visible in the display, identify the author whose work stayed on \textit{The New York Times} Bestseller List for the most total weeks during the years when ``Authors United'' was most active. Report the author, the work, and the total weeks.

\par\vspace{1mm}
\textit{Input image.}
\begin{center}
  \includegraphics[height=3.4cm, keepaspectratio]{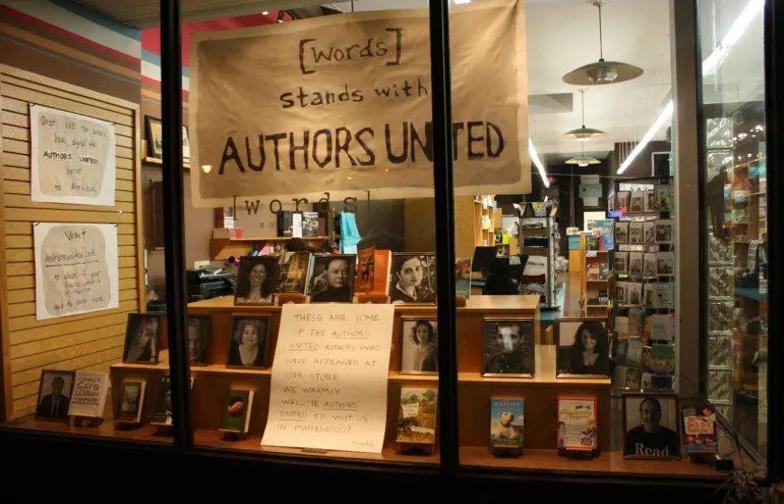}
\end{center}

\vspace{1mm}
\textbf{Ground truth.} Donna Tartt; \textit{The Goldfinch}; 39 weeks.
\vspace{5pt}
\par\hrule\vspace{2mm}
\textbf{Tool trajectory (condensed).}

\textbf{Step 1.} \textit{web search} \\
\textbf{Goal.} Identify the bookstore and locate a source describing the window display. \\
\textbf{Outcome.} Finds a relevant local article about the display.

\textbf{Step 2.} \textit{visit} \\
\textbf{Goal.} Read the article and recover the set of authors shown in the display. \\
\textbf{Outcome.} The page provides context but does not yield a complete, reliable list of the visible authors.

\textbf{Step 3.} \textit{code interpreter} \\
\textbf{Goal.} Extract author names from the image via OCR. \\
\textbf{Outcome.} OCR is noisy and incomplete; text cues are not recovered reliably.

\textbf{Step 4.} \textit{code interpreter} \\
\textbf{Goal.} Crop the window display into smaller regions to identify individual authors. \\
\textbf{Outcome.} Produces multiple crops, but faces and cover details remain ambiguous at this resolution.
\begin{center}
  \includegraphics[height=3.0cm, keepaspectratio]{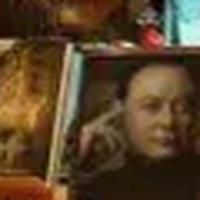}\hspace{1mm}
  \includegraphics[height=3.0cm, keepaspectratio]{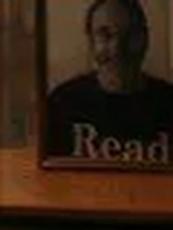}\hspace{1mm}
  \includegraphics[height=3.0cm, keepaspectratio]{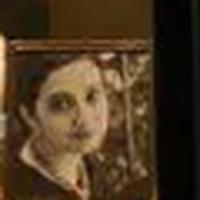}\hspace{1mm}
  \includegraphics[height=3.0cm, keepaspectratio]{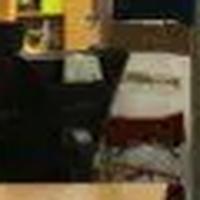}\\[0.8mm]
  \includegraphics[height=3.0cm, keepaspectratio]{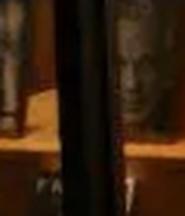}\hspace{1mm}
  \includegraphics[height=3.0cm, keepaspectratio]{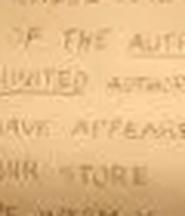}\hspace{1mm}
  \includegraphics[height=3.0cm, keepaspectratio]{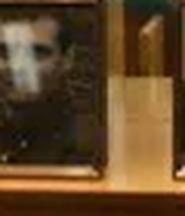}\hspace{1mm}
  \includegraphics[height=3.0cm, keepaspectratio]{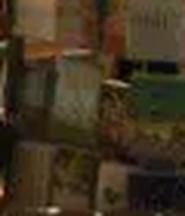}\\[0.8mm]
  \includegraphics[height=3.0cm, keepaspectratio]{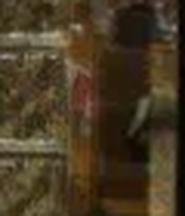}\hspace{1mm}
  \includegraphics[height=3.0cm, keepaspectratio]{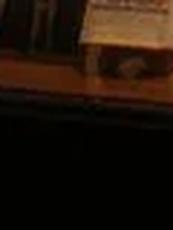}
\end{center}

\textbf{Steps 5--9.} \textit{image search / web search / visit} \\
\textbf{Goal.} Identify cropped portraits and posters (e.g., ``READ'' poster) via reverse search and related pages. \\
\textbf{Outcome.} Reverse search does not return actionable matches; the model cannot reliably map crops to specific authors.

\textbf{Steps 10--13.} \textit{web search} \\
\textbf{Goal.} Estimate the correct answer by comparing bestseller-list durations among guessed candidates. \\
\textbf{Outcome.} The model anchors on the wrong author set and proceeds with an incorrect comparison.

\vspace{5pt}
\par\hrule\vspace{2mm}

\textbf{Failure point.} \textbf{Step 4 (visual identification).} Even after cropping, the model fails to correctly identify the clearly visible author (Donna Tartt) from the display. This incorrect visual grounding leads to downstream searches and bestseller comparisons over the wrong candidate set, culminating in an incorrect final answer.
\vspace{5pt}
\par\hrule\vspace{2mm}
\textbf{Final answer.} Anthony Doerr; \textit{All the Light We Cannot See}; $\sim$84 weeks. \No

\end{tcolorbox}

\begin{tcolorbox}[breakable,
  colframe=dustyrose,
  colback=white,
  coltitle=white,
  title={Traj \#5: Target Arena Identification (\textsc{Gemini-3-Pro}; failed; 11 tool calls)},
  fonttitle=\bfseries,
  before skip=5pt]
\small

\textbf{Task.} A colleague referenced this venue only as ``Target Arena A.'' Identify exactly which university facility is shown so the correct team metadata (e.g., seating capacity, fan distribution) can be linked in a database.

\par\vspace{1mm}
\textit{Input image.}
\begin{center}
    \includegraphics[height=3.4cm, keepaspectratio]{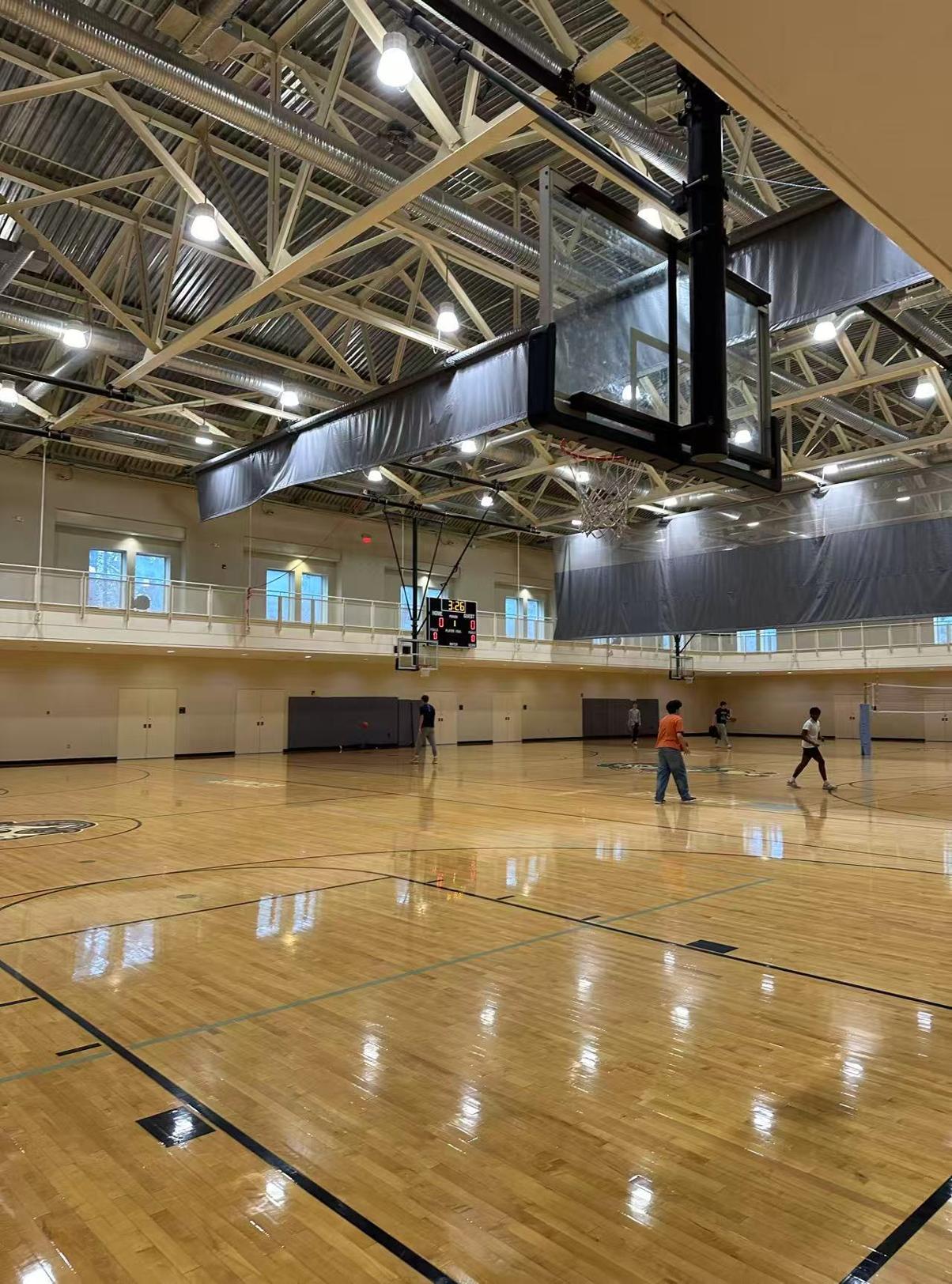}
\end{center}

\vspace{1mm}
\textbf{Ground truth.} UNC
\vspace{5pt}
\par\hrule\vspace{2mm}
\textbf{Tool trajectory (condensed).}

\textbf{Step 1.} \textit{code interpreter} \\
\textbf{Goal.} Inspect the full scene and locate the most informative identifiers (court logo, scoreboard, signage). \\
\textbf{Outcome.} The model focuses on the floor logo and plans to use it for identification.

\textbf{Step 2.} \textit{code interpreter} \\
\textbf{Goal.} Crop the court logo and scoreboard region for downstream search. \\
\textbf{Outcome.} Produces a small, low-detail logo crop (insufficient for reliable matching).
\begin{center}
  \includegraphics[height=3.0cm, keepaspectratio]{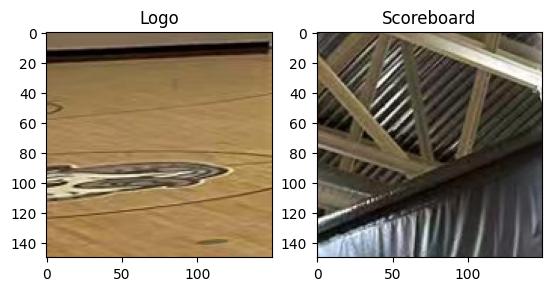}
\end{center}

\textbf{Step 3.} \textit{image search (reverse)} \\
\textbf{Goal.} Identify the venue or team via reverse search on the cropped logo. \\
\textbf{Outcome.} No matches returned.

\textbf{Step 4.} \textit{code interpreter} \\
\textbf{Goal.} Re-crop the logo with more context. \\
\textbf{Outcome.} The crop remains ambiguous and visually distorted.
\begin{center}
  \includegraphics[height=3.0cm, keepaspectratio]{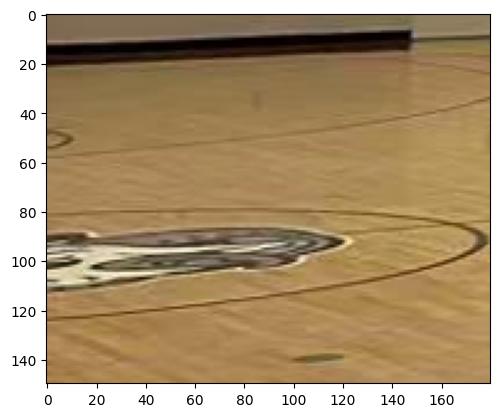}
\end{center}

\textbf{Step 5.} \textit{image search (reverse)} \\
\textbf{Goal.} Retry reverse search with the updated crop. \\
\textbf{Outcome.} Still no matches.

\textbf{Step 6.} \textit{web search} \\
\textbf{Goal.} Use structural cues (exposed truss ceiling, running track/balcony) to narrow candidates. \\
\textbf{Outcome.} Returns generic lists; does not provide a grounded identification.

\textbf{Step 7.} \textit{code interpreter} \\
\textbf{Goal.} Enhance contrast/sharpness of the logo crop to clarify the mascot shape. \\
\textbf{Outcome.} Enhancement does not resolve the mascot identity.
\begin{center}
  \includegraphics[height=3.0cm, keepaspectratio]{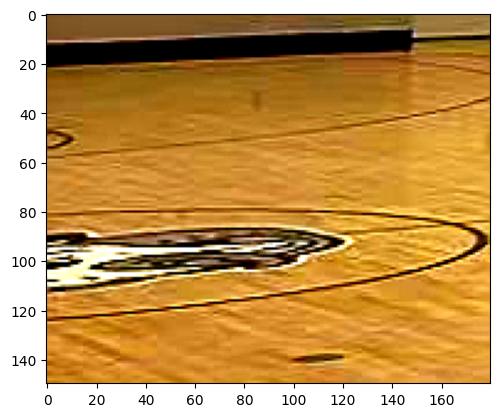}
\end{center}

\textbf{Step 8.} \textit{image search (reverse)} \\
\textbf{Goal.} Reverse search the enhanced crop. \\
\textbf{Outcome.} No matches.

\textbf{Step 9.} \textit{image search (text)} \\
\textbf{Goal.} Pivot to a guessed candidate (University of St.\ Thomas AARC) based on weak visual similarity. \\
\textbf{Outcome.} Retrieves reference images for comparison.

\textbf{Step 10.} \textit{image search (text)} \\
\textbf{Goal.} Expand references (Schoenecker Arena / AARC interiors) and match architecture. \\
\textbf{Outcome.} The model over-weights generic ceiling/track similarities.

\textbf{Step 11.} \textit{image search (text)} \\
\textbf{Goal.} Final confirmation via additional reference images. \\
\textbf{Outcome.} Commits to the incorrect university.

\vspace{5pt}
\par\hrule\vspace{2mm}
\textbf{Failure point.} \textbf{Step 9 (incorrect anchoring).} After failing to identify the floor logo, the model switches to architecture-based matching and prematurely anchors on St.\ Thomas. The remaining steps reinforce this guess using generic similarities (trusses, ducts, balcony/track) rather than a definitive visual identifier from the query image, leading to an incorrect final answer.
\vspace{5pt}
\par\hrule\vspace{2mm}
\textbf{Final answer.} University of St.\ Thomas (Minnesota) \No

\end{tcolorbox}

\begin{tcolorbox}[breakable,
  colframe=dustyrose,
  colback=white,
  coltitle=white,
  title={Traj \#6: Pilea Root Diagnosis (\textsc{Gemini-3-Pro}; failed; 6 tool calls)},
  fonttitle=\bfseries]
\small

\textbf{Task.} A Pilea was overwatered two months ago and later stabilized. During repotting, a large white mass is found at the root base; it is hard and crusty (does not burst when poked). Based on the images and history, identify the substance and decide whether pesticides are needed.

\par\vspace{1mm}
\textit{Input images.}
\begin{center}
    \includegraphics[height=3.4cm, keepaspectratio]{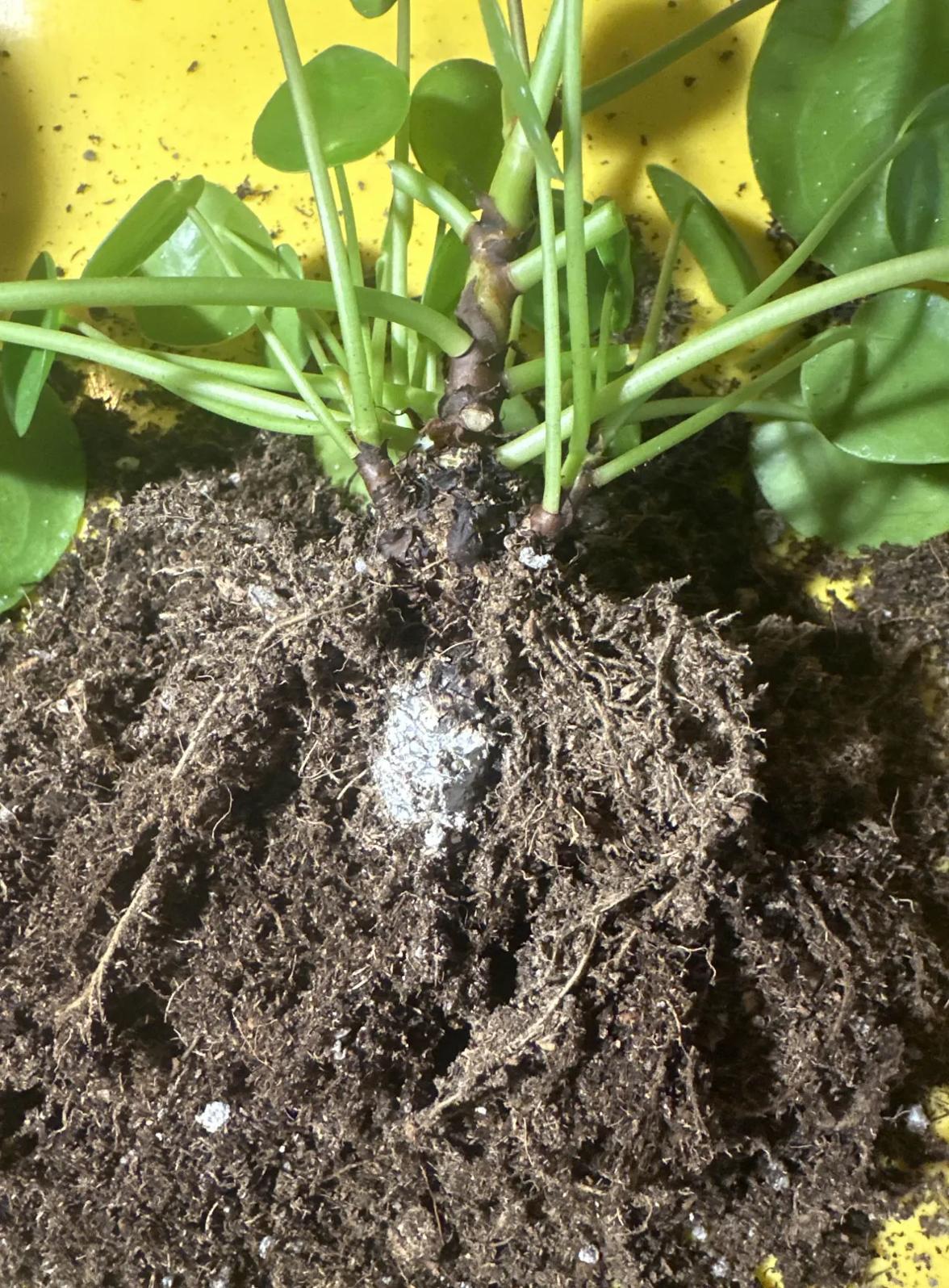} \hspace{4mm}
    \includegraphics[height=3.4cm, keepaspectratio]{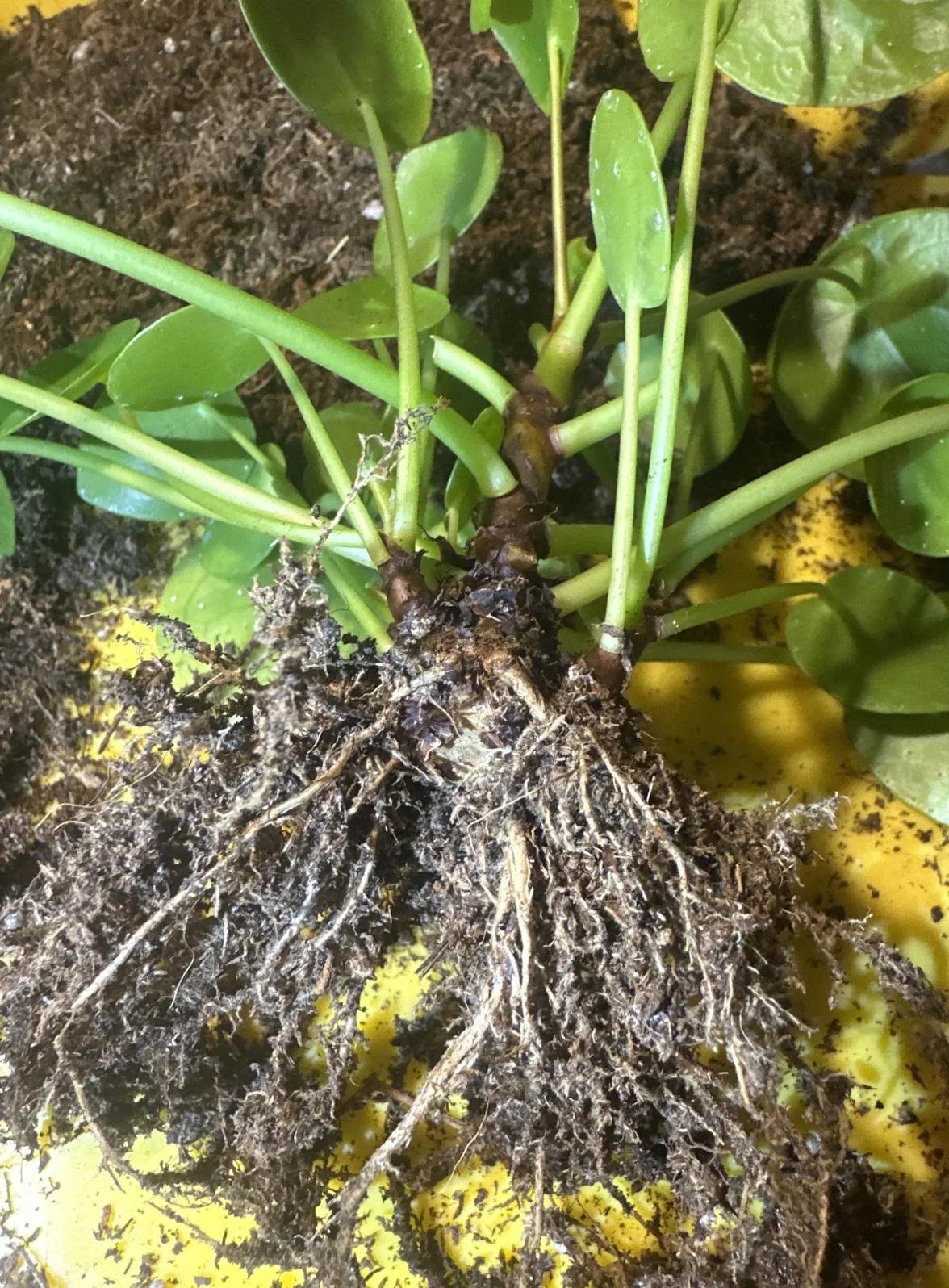} \hspace{4mm}
    \includegraphics[height=3.4cm, keepaspectratio]{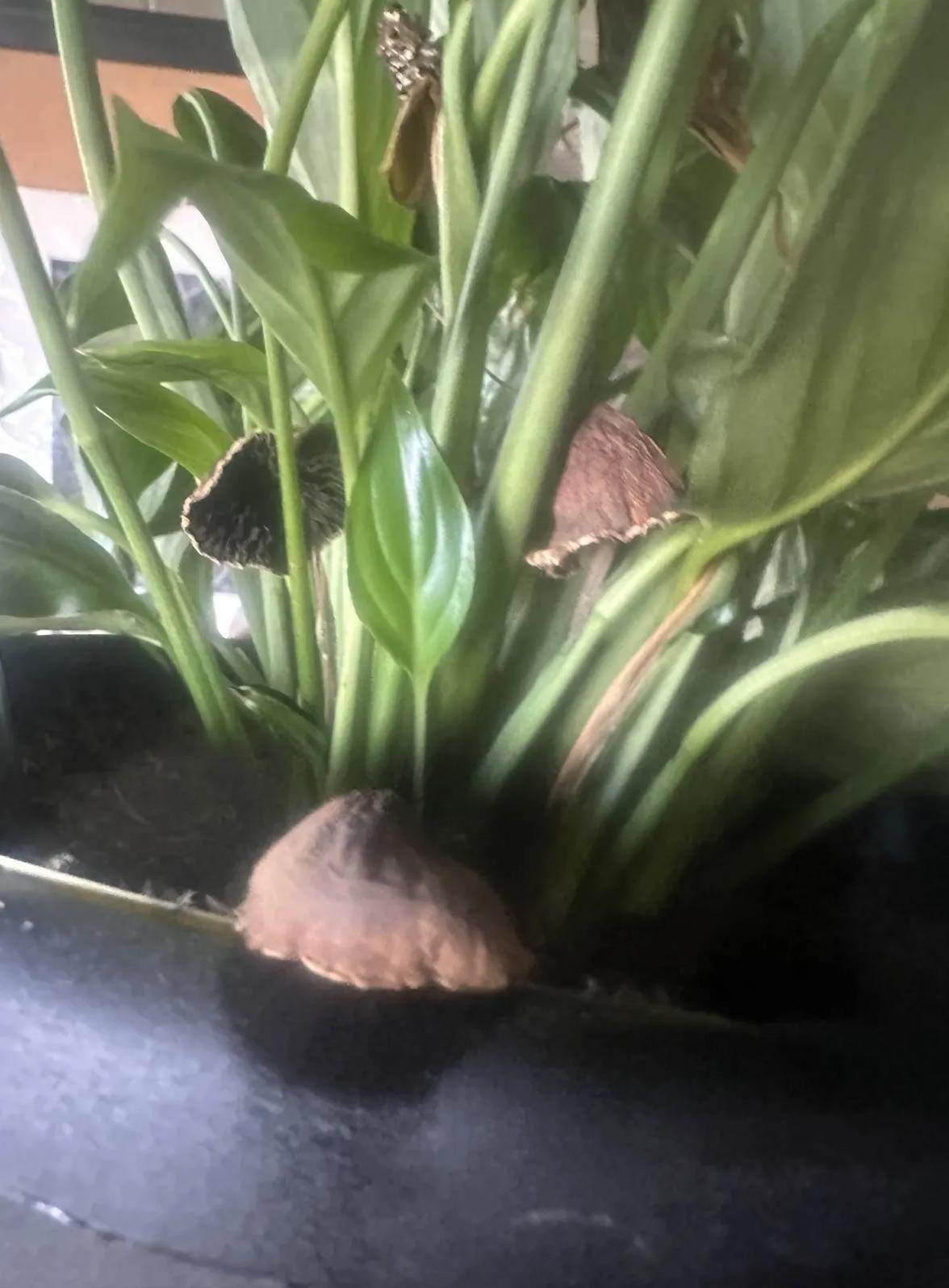}
\end{center}

\vspace{1mm}
\textbf{Ground truth.} No pesticide needed. The mass is calloused / crustified root-rot residue from the prior overwatering incident.
\vspace{5pt}
\par\hrule\vspace{2mm}
\textbf{Tool trajectory (condensed).}

\textbf{Step 1.} \textit{web search} \\
\textbf{Goal.} Identify common causes of a hard white mass near Pilea roots. \\
\textbf{Outcome.} Results skew toward unrelated ``white spots'' / mineral topics (mostly leaf-focused).

\textbf{Step 2.} \textit{web search} \\
\textbf{Goal.} Broaden to generic houseplant root ``white lump'' explanations. \\
\textbf{Outcome.} Mixed hypotheses (mineral buildup, mold/fungus) without matching the specific context.

\textbf{Step 3.} \textit{image search} \\
\textbf{Goal.} Look up nursery-plug examples as an alternative explanation. \\
\textbf{Outcome.} Retrieves plug images that encourage an external ``plug/mesh'' hypothesis.

\textbf{Step 4.} \textit{web search} \\
\textbf{Goal.} Explore other possibilities (e.g., slime mold / residue) from search cues. \\
\textbf{Outcome.} Finds generic statements that crusty residue can occur, but without strong alignment to an underground root-base mass.

\textbf{Step 5.} \textit{web search} \\
\textbf{Goal.} Re-search with Pilea-specific phrasing to confirm the plug/mesh theory. \\
\textbf{Outcome.} Surfacing discussions about ``root mesh / death plugs'' further reinforces the plug narrative.

\textbf{Step 6.} \textit{web search} \\
\textbf{Goal.} Finalize between plug/mesh vs.\ plant-tissue explanation. \\
\textbf{Outcome.} Commits to ``nursery plug + dried fungus'' as the main explanation.

\vspace{1mm}
\vspace{5pt}
\par\hrule\vspace{2mm}

\textbf{Failure point.} \textbf{Step 3 (incorrect hypothesis shift).} After early searches returned noisy, non-root-specific results, the model pivots to an external ``nursery plug / mesh'' explanation based on loosely similar reference images. This guess then becomes the anchor and is reinforced by subsequent searches, while the overwatering history and root-location cue (consistent with healed root damage) are underweighted.
\vspace{5pt}
\par\hrule\vspace{2mm}
\textbf{Final answer.} A nursery plug (``death plug'') encrusted/colonized by dried saprophytic fungus; no pesticides. \No

\end{tcolorbox}

\begin{tcolorbox}[breakable,
  colframe=dustyrose,
  colback=white,
  coltitle=white,
  title={Traj \#7: Swing Prop Construction (\textsc{Gemini-3-Pro}; direct response)},
  fonttitle=\bfseries]
\small

\textbf{Task.} Build a swing prop for a children's photography studio that \emph{looks} like a rope-hung swing but is completely stationary and safe. The seat must appear connected to the top beam (no ``floating'' look). Only basic woodworking tools are available.

\par\vspace{1mm}
\textit{Input image.}
\begin{center}
    \includegraphics[height=3.4cm, keepaspectratio]{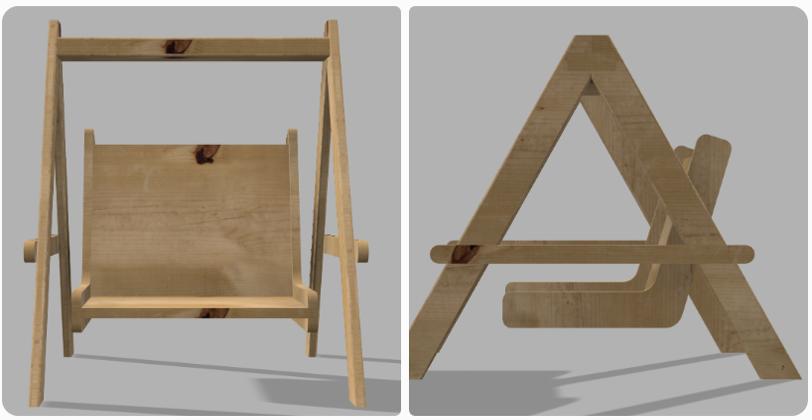}
\end{center}

\vspace{1mm}
\textbf{Ground truth.} Bolt the seat to the frame using vertical wooden posts, then wrap thick hemp rope around the posts to hide the wood and create the hanging illusion.
\vspace{5pt}
\par\hrule\vspace{2mm}
\textbf{Model response (condensed).}

\textbf{Step 1 (direct).}
\vspace{-5pt}
\begin{itemize}[leftmargin=*]
    \setlength{\itemsep}{2pt}
    \item \textbf{Interpretation.} The model treats the prop as a bench-like structure supported by a visible horizontal stretcher bar, rather than creating a concealed rigid support that still looks rope-hung.
    \item \textbf{Proposed design.} Build a rigid A-frame, add a horizontal stretcher across the legs, mount the seat on the stretcher, and add decorative (non-load-bearing) ropes from top beam to seat to mimic a swing.
\end{itemize}

\par\hrule\vspace{2mm}

\textbf{Failure point.} \textbf{Constraint miss (visual illusion).} The proposed stretcher-bar support remains visually apparent and undermines the requirement that the seat should look convincingly hung from the top beam. It does not use hidden vertical supports wrapped in rope, which is the key trick in the ground-truth solution.
\vspace{5pt}
\par\hrule\vspace{2mm}
\textbf{Final answer.} Build a stationary bench supported by a horizontal bar, then add taut decorative ropes to imitate a swing. \No

\end{tcolorbox}


\end{document}

%% file: table/comparison_datasets.tex
\begin{table}[t]
\centering
\small
\setlength{\tabcolsep}{4pt}
\renewcommand{\arraystretch}{1.05}
\caption{Comparison with representative multimodal agent benchmarks. Operation abbreviations: \textbf{\textsc{VO.}} (Visual Operations), \textbf{\textsc{VS.}} (Visual Search), \textbf{\textsc{TS.}} (Text Search), and \textbf{\textsc{CE.}} (Code Execution). Tool categories are based on the tools and signals used in these benchmarks. ``\# Turns'' reports the average number of tool-calling turns by \textsc{GPT-5}, used as a proxy for task complexity.}

\label{tab:agent_benchmark_comparison}

\resizebox{1.0\linewidth}{!}{%
\begin{tabular}{l|cccccc}
\toprule
\textbf{\textsc{Benchmark}} &
\textbf{\textsc{VO.}} & \textbf{\textsc{VS.}} & \textbf{\textsc{TS.}} & \textbf{\textsc{CE.}} &
\textbf{\makecell{\textsc{Multi}\\\textsc{Image}}} &
\textbf{\makecell{\# Turns}}\\
\midrule
\textsc{TIR-Bench}       & \Yes & \No  & \No  & \Yes  & \Yes & 2.92 \\
\textsc{Agent-X}         & \Yes
& \No  & \Yes  & \Yes  & \Yes & 3.4  \\
\textsc{MMSearch-Plus}   & \No  & \Yes & \Yes & \No & \Yes  & 4.6 \\
\textsc{BrowseComp-VL}   & \No  & \Yes & \Yes & \Yes & \No  & 4.3 \\
\textsc{VisualToolBench} & \Yes & \No  & \Yes & \Yes  & \No & 4.46  \\
\midrule
\textsc{AgentVista} (Ours) & \Yes & \Yes & \Yes & \Yes & \Yes & 12.67\\
\bottomrule
\end{tabular}%
}
\vspace{-10pt}
\end{table}

%% file: table/statistics.tex
\begin{table}[t]
    \centering
    \caption{Summary statistics of the \textsc{AgentVista} benchmark.}
    \small
    \setlength{\tabcolsep}{6pt} 
    \begin{tabular}{p{5.5cm} r}
        \toprule
        \textsc{\textbf{Statistic}} & \textsc{\textbf{Number}} \\
        \midrule
        Total queries & 209 \\
        Total images & 308 \\
        Primary categories & 7 \\
        Secondary categories & 25 \\
        \midrule
        Average query length & 401.4 \\
        Average answer length & 40.8 \\
        \midrule
        Image distribution & \\
        \hspace{0.3cm}- Single-image queries & 151 (72.2\%) \\
        \hspace{0.3cm}- Multi-image queries & 58 (27.8\%) \\
        
        \bottomrule
    \end{tabular}
    \vspace{-10pt}
    \label{tab:statistics}
\end{table}

%% file: table/main_table.tex
\begin{table*}[t!]
  \caption{Main results on our proposed \textsc{AgentVista}.
  Domain abbreviations:
  \textbf{\textsc{Comm.}} (Commerce), \textbf{\textsc{Geog.}} (Geography), \textbf{\textsc{Ent.}} (Entertainment),
  \textbf{\textsc{Tech.}} (Technology), \textbf{\textsc{Soc.}} (Society), \textbf{\textsc{Acad.}} (Academics),
  and \textbf{\textsc{Cult.}} (Culture).
  Input mode abbreviations: \textbf{\textsc{Single.}} (Single-image input) and \textbf{\textsc{Multi.}} (Multi-image input).
  The best-performing model in each category is \textbf{in-bold}, and the second best is \underline{underlined}. Overall, \textsc{Gemini-3-Pro} achieves the highest accuracy among all evaluated models. All values are accuracies in \%.}
  \vspace{-5pt}
  \label{main-table}
  \centering
   \resizebox{1.0\linewidth}{!}{
    \begin{tabular}{l|ccccccc|cc|cc}
      \toprule
      \multirow{2}{*}{\textbf{\textsc{Model}}}
      & \multicolumn{7}{c|}{\textbf{\textsc{By Category}}}
      & \multicolumn{2}{c|}{\textbf{\textsc{By Input Mode}}}
      & \multicolumn{2}{c}{\textbf{\textsc{Summary}}} \\
      \cmidrule(lr){2-8} \cmidrule(lr){9-10} \cmidrule(lr){11-12}
      & \textbf{\textsc{Comm.}} & \textbf{\textsc{Geog.}} & \textbf{\textsc{Ent.}} & \textbf{\textsc{Tech.}} & \textbf{\textsc{Soc.}} & \textbf{\textsc{Acad.}} & \textbf{\textsc{Cult.}}
      & \textbf{\textsc{Single.}} & \textbf{\textsc{Multi.}}
      & \textbf{\textsc{Overall}} & \textbf{\textsc{\# Turns}} \\
      \midrule
      \textsc{Qwen3-VL-235B} & 7.14 & 7.69 & 7.69 & 26.47 & 16.00 & 20.00 & 13.33 & 11.84 & 15.79 & 12.92 & 2.34 \\
      \textsc{GPT-4.1} & 16.67 & 15.38 & 10.26 & 29.41 & 20.00 & 20.00 & 13.33 & 15.13 & 24.56 & 17.70 & 1.74 \\
      \textsc{o3} & \underline{21.43} & 15.38 & 7.69 & 23.53 & \textbf{40.00} & 26.67 & 13.33 & 17.76 & 26.32 & 20.10 & 13.18 \\
      \textsc{o4-mini} & 2.38 & 10.26 & 2.56 & 8.82 & 8.00 & 13.33 & 0.00 & 6.58 & 5.26 & 6.22 & 1.89 \\
      \textsc{GPT-5} & \textbf{23.81} & \underline{23.08} & 12.82 & \underline{35.29} & 28.00 & 26.67 & \underline{26.67} & \textbf{24.34} & 24.56 & \underline{24.40} & 12.67 \\
      \textsc{GPT-5.1} & \textbf{23.81} & 12.82 & \underline{15.38} & 26.47 & 24.00 & \textbf{40.00} & \textbf{40.00} & 19.74 & \underline{31.58} & 22.97 & 17.14 \\
      \textsc{GPT-5.2} & \underline{21.43} & 17.95 & \textbf{20.51} & \textbf{38.24} & 24.00 & \underline{33.33} & 20.00 & 23.03 & 28.07 & \underline{24.40} & 13.85 \\
      \textsc{Grok-4} & 11.90 & \underline{23.08} & 7.69 & 20.59 & 28.00 & 0.00 & 0.00 & 13.82 & 17.54 & 14.83 & 16.44 \\
      \textsc{Claude-Sonnet-4} & 9.52 & 15.38 & 2.56 & 29.41 & 16.00 & 20.00 & 6.67 & 11.18 & 21.05 & 13.88 & 5.37 \\
      \textsc{Claude-Opus-4} & 19.05 & 12.82 & 5.13 & 26.47 & 20.00 & 20.00 & 6.67 & 11.84 & 26.32 & 15.79 & 6.89 \\
      \textsc{Claude-Opus-4.1} & 11.90 & \underline{23.08} & 10.26 & 29.41 & 16.00 & 26.67 & 13.33 & 16.45 & 22.81 & 18.18 & 7.28 \\
      \textsc{Claude-Sonnet-4.5} & 11.90 & \underline{23.08} & 7.69 & 26.47 & 24.00 & 20.00 & 13.33 & 17.11 & 19.30 & 17.70 & 9.99 \\
      \textsc{Gemini-3-Flash} & 16.67 & 17.95 & 10.26 & 29.41 & 28.00 & \textbf{40.00} & 20.00 & 18.42 & 28.07 & 21.05 & 7.78 \\
      \textsc{Gemini-3-Pro} & 16.67 & \textbf{28.21} & \textbf{20.51} & 32.35 & \underline{32.00} & \textbf{40.00} & \textbf{40.00} & \underline{23.68} & \textbf{36.84} & \textbf{27.27} & 6.67 \\
      \bottomrule
    \end{tabular}
    
    }
    \vspace{-10pt}
\end{table*}